\documentclass{article}
\usepackage{iclr2026_conference,times}
\usepackage{hyperref}
\usepackage{url}

\usepackage{graphicx}
\usepackage{amsmath,amssymb,bm,mathtools}
\usepackage{booktabs,multirow,array}
\usepackage{microtype}
\usepackage{xcolor}
\usepackage{pifont}
\usepackage{xspace}
\usepackage{tikz}
\usepackage{makecell}
\usepackage{xurl}
\usetikzlibrary{shapes.geometric,arrows.meta,positioning,fit,
                backgrounds,calc,patterns}
\definecolor{tblue}{RGB}{21,101,192}
\definecolor{torange}{RGB}{230,81,0}
\definecolor{tgold}{RGB}{180,120,0}
\definecolor{tgreen}{RGB}{46,125,50}
\definecolor{tgray}{RGB}{100,100,100}
\definecolor{tred}{RGB}{183,28,28}
\tikzset{
  resblk/.style={rectangle,rounded corners=2pt,draw=#1,fill=#1!12,
    minimum width=19mm,minimum height=5.5mm,font=\scriptsize,inner sep=2pt},
  attnblk/.style={rectangle,rounded corners=2pt,draw=tgold,fill=tgold!18,
    minimum width=19mm,minimum height=5.5mm,font=\scriptsize,inner sep=2pt},
  modbox/.style={rectangle,rounded corners=4pt,draw=#1,thick,
    minimum width=28mm,font=\small\bfseries,inner sep=4pt},
  cha/.style={font=\tiny,text=#1},
  nar/.style={-Stealth,#1,semithick},
  los/.style={-Stealth,very thick,#1},
}
\providecommand{\ie}{}\renewcommand{\ie}{\textit{i.e.},\xspace}

\providecommand{\etal}{}\renewcommand{\etal}{\textit{et al.}\xspace}
\newcommand{\dash}{\textsc{DASH}\xspace}
\newcommand{\cmark}{\ding{51}}
\newcommand{\xmark}{\ding{55}}
\newcommand{\snr}{\mathrm{SNR}}
\newcommand{\eps}{\ensuremath{\boldsymbol{\varepsilon}}}
\newcommand{\xt}{\ensuremath{\mathbf{x}_t}}
\newcommand{\xo}{\ensuremath{\mathbf{x}_0}}
\newcommand{\Lim}{\ensuremath{\mathcal{L}_{\mathrm{im}}}\xspace}
\newcommand{\Lun}{\ensuremath{\mathcal{L}_{\mathrm{un}}}\xspace}
\newcommand{\Lan}{\ensuremath{\mathcal{L}_{\mathrm{an}}}\xspace}
\newcommand{\Ltot}{\ensuremath{\mathcal{L}_{\mathrm{DASH}}}\xspace}

\newcommand{\figorbox}[2]{%
  \IfFileExists{#2}%
    {\includegraphics[width=#1]{#2}}%
    {\fcolorbox{gray!55}{gray!8}{%
       \begin{minipage}[c][0.60\dimexpr#1\relax][c]%
                       {\dimexpr#1-2\fboxsep-2\fboxrule\relax}
         \centering\ttfamily\tiny\color{gray!75}
         [missing]\\\detokenize{#2}
       \end{minipage}}}%
}
\title{DASH: Dual-Branch Score Distillation for\\
       Guidance-Calibrated Compact Diffusion Models}

\author{%
  Abdullah Al Shafi\textsuperscript{1},
  Kazi Saeed Alam\textsuperscript{1},
  Sk Imran Hossain\textsuperscript{1},
  Engelbert Mephu Nguifo\textsuperscript{2} \\
  \textsuperscript{1}Khulna University of Engineering \& Technology, Bangladesh \\
  \textsuperscript{2}University Clermont Auvergne, France \\
  \texttt{abdullah.shafi99@gmail.com} \\
  \texttt{\{saeed.alam, imran\}@cse.kuet.ac.bd} \\
  \texttt{engelbert.mephu\_nguifo@uca.fr}
}

\iclrfinalcopy

\begin{document}
\maketitle

\begin{abstract}
Parameter compression of class-conditional diffusion models exposes a
structural limitation in output-level distillation: supervising only
the guided output leaves the two score branches non-identifiable, so
the classifier-free guidance gap is not determined in the student. The
zero-loss set admits degenerate solutions in which both branches
collapse toward identical predictions, and guidance loses its effect at
inference despite low training loss. For each fixed input the objective is exactly flat in a branch-output
direction at every residual value, so the ambiguity is a property of
the output-level objective rather than a poor local minimum. This paper introduces \dash, which
supervises the conditional and unconditional branches independently.
An anchor term regularises the conditional prediction toward
ground-truth noise, and the teacher's final learned per-timestep curriculum
transfers into the student as a frozen prior. Across CIFAR-10,
CIFAR-100, and class-conditional ImageNet-64, a more than
$5{\times}$ compressed student stays within four FID points of its
teacher and recovers at least $89\%$ of the teacher's guidance-gap
magnitude, where no other two-branch baseline with defined
calibration exceeds $82\%$. Ablation
isolates unconditional supervision as the term that separates this
formulation from output-level distillation. The account is
quantitative: the null space fixes the ratio between a composite
student's two branch errors, and the measured ratios approximately
match that prediction on all three datasets.
\end{abstract}

\section{Introduction}
\label{sec:intro}

Denoising diffusion probabilistic models~\cite{ddpm} have established
themselves as the dominant framework for high-quality image
synthesis~\cite{adm,edm}, with classifier-free guidance
(CFG)~\cite{cfg} enabling precise conditional control without auxiliary
classifiers. CFG trains a single network to produce both conditional
($\boldsymbol{\varepsilon}^y$) and unconditional
($\boldsymbol{\varepsilon}^\varnothing$) predictions through stochastic
label dropout, combining them at inference as
$\tilde{\boldsymbol{\varepsilon}} = \boldsymbol{\varepsilon}^\varnothing
+ w(\boldsymbol{\varepsilon}^y - \boldsymbol{\varepsilon}^\varnothing)$,
where the guidance gap
$\Delta = \boldsymbol{\varepsilon}^y - \boldsymbol{\varepsilon}^\varnothing$
is amplified at every denoising step to steer the generation trajectory.
The quality of this amplification depends critically on both the
magnitude and direction of $\Delta$: an undercalibrated gap suppresses
class-specific sharpening, while a collapsed gap renders the guidance
weight $w$ entirely ineffective. Despite these benefits, CFG doubles
per-step computational cost and requires substantial model capacity.
Deploying it in resource-constrained settings therefore depends on
parameter compression that preserves the calibrated guidance mechanism.

Compressing CFG-guided diffusion models exposes a property of
output-level distillation objectives: supervising only the composite
guided prediction imposes a single constraint on two independent
score branches, admitting degenerate solutions where the guidance
gap collapses ($\Delta_S{\to}0$) even when the training objective is
well-optimised (\S\ref{sec:method}). Collapse is not a saddle point
or an initialisation artefact: it is a global minimiser of the
composite objective. This training-time consequence is obscured by
CFG's treatment as a purely inference-time mechanism;
\S\ref{sec:related} situates it against theoretical analyses of
composite objectives that stop short of this branch-level
characterisation.

\dash resolves this non-identifiability by supervising each branch
directly. The unconditional loss
$\mathcal{L}_\mathrm{un}$ directly constrains
$\boldsymbol{\varepsilon}^\varnothing_S \approx
\boldsymbol{\varepsilon}^\varnothing_T$ and the imitation loss
$\mathcal{L}_\mathrm{im}$ matches
$\boldsymbol{\varepsilon}^y_S \approx \boldsymbol{\varepsilon}^y_T$,
together forming a fully determined joint objective whose unique
minimiser matches both teacher branches simultaneously. A lightweight
anchor term $\mathcal{L}_\mathrm{an}$ regularises the conditional
branch toward ground-truth noise, preventing drift during early
training when the teacher target provides a weak signal.

Ablation confirms the mechanism
rather than only the cost: removing $\mathcal{L}_\mathrm{un}$ drives
$\rho$ from $0.91$ to $0.23$, and composite residuals sit on the null
line at the ratio $(w{-}1)/w$ it forces
(\S\ref{sec:distill_results}). Eight
alternative supervision and compression formulations (composite,
multi-$w$ composite, conditional-only, gap matching, label-dropout,
FitNets, magnitude pruning, and single-head training) either recover
the guidance gap incompletely or discard the branch decomposition, and
training from scratch does not reach the teacher at all; dual-branch
supervision is the only formulation tested that recovers the gap to
$\rho{>}0.85$ on both datasets.

Beyond resolving non-identifiability, \dash transfers the teacher's
final learned per-timestep training curriculum to the student as a
frozen prior (TIRT, Timestep-Importance Rebalanced Training;
Transfer described in \S\ref{sec:distil}), since teacher and student
error profiles need not drive the schedule to the same place. Substituting a fixed Min-SNR schedule for this
transfer costs 2.60\,/\,2.67 FID on the two datasets
(Table~\ref{tab:ablation}). The transfer is applied
identically to every distilled loss-formulation baseline, so it does not contribute to
the margins reported between loss formulations.

Experiments on CIFAR-10 and CIFAR-100 validate the framework at two
levels of class granularity. At $5.9{\times}$ parameter compression the
student stays within 4 FID points of its teacher, with
$\rho{=}0.91$ guidance calibration. On CIFAR-100 \dash reaches FID~10.47 against composite
distillation's 20.84, a wider margin than the 8.87 against 13.84 on
CIFAR-10. The same non-identifiability and its correction carry to
class-conditional ImageNet-64 (Table~\ref{tab:in64}), where composite
distillation reaches only $\rho{=}0.65$ against \dash's $0.89$,
the same deficit composite shows at 100 classes. The mechanism is
therefore not confined to $32{\times}32$ inputs or 10/100-class label spaces.

\section{Related Work}
\label{sec:related}

\noindent\textbf{Diffusion models and classifier-free guidance.}
Denoising diffusion models~\cite{ddpm,iddpm,adm,edm} establish noise
prediction as the dominant generative paradigm; classifier-free
guidance forms each denoising step by combining a conditional and an
unconditional score. Recent work refines that combination at inference,
constraining score updates to the data manifold~\cite{cfgpp}, applying
guidance only to mid-trajectory timesteps~\cite{intervalcfg}, or
substituting a smaller, less-trained model for the unconditional
branch~\cite{autoguidance}. None retrains the branches themselves under
reduced capacity, so none addresses calibration when the deployed
student is compressed; \dash instead supervises both branches of a
single compressed student during training, rather than substituting
or reweighting either at inference.

\noindent\textbf{Step reduction and parameter compression.}
Step-reduction
distillation~\cite{luhman2021,progdistill,cm,ctm,lcm,add,dmd} and
solver-based sampling~\cite{dpmsolver,dpm_plus} attain few-step
sampling at teacher-scale capacity, and
guidance-specific variants remove CFG's second forward pass rather than
parameters, either by distilling the composite output into a single
$w$-conditioned student~\cite{meng2023distill} or through adapters on a
frozen base model~\cite{agd2025}. Parameter compression shrinks the
network instead, but existing methods leave the CFG branches
unconstrained: BK-SDM~\cite{bksdm} and SnapFusion~\cite{snapfusion}
target latent-space text-conditional models, BOOT~\cite{boot}
preserves teacher capacity, O2MKD~\cite{o2mkd} compresses
unconditional models exposing no branch decomposition,
and Diff-Pruning~\cite{diffpruning} prunes a trained teacher using
Taylor-based importance; the compression baseline here adopts the same
prune-then-finetune structure but ranks filters by magnitude
(\S\ref{sec:impl}). The two axes
are orthogonal: \dash reduces the student to 6.1M parameters at the
full 50-step budget and is composable with step reduction.

\noindent\textbf{Timestep weighting and curriculum transfer.}
Min-SNR~\cite{minsnr} reweights timesteps from the noise schedule and
perception-prioritised training~\cite{choi2022perception} from
perceptual rankings; both schedules are fixed and data-agnostic.
EDM2~\cite{edm2} instead learns a per-noise-level weighting jointly
with the network, as TIRT does. A
separate line reweights along the \emph{sample}
axis~\cite{doremi,infobatch,regmix,li2025pruning,kim2024unbiased},
selecting which examples to prioritise rather than the timestep axis
TIRT acts on. Transferring such a \emph{final learned} weighting from
teacher to student (\S\ref{sec:distil}), rather than relearning it,
has not, to the authors' knowledge, been explored in prior
distillation work.

\noindent\textbf{Branch-level supervision and identifiability.}
Supervision need not act on outputs at all: FitNets~\cite{fitnets}, as originally
proposed, aligns intermediate feature maps, constraining what the
student computes internally but not the score branches it emits. A theoretical treatment of the composite guided
objective~\cite{distilldirect} stops short of the branch-level
characterisation given here. A distinct line of work
\emph{combines} the two branches into a single forward pass to
accelerate sampling~\cite{meng2023distill}; \dash instead
\emph{separates} them under an additional constraint to preserve
calibration under capacity reduction, the opposite operation toward an
orthogonal goal. Concurrent work~\cite{li2026rethinking} identifies a
related branch-level under-identification in on-policy CFG
distillation, arising dynamically rather than from the static
null-space characterisation proven here.
Across the closest prior methods (step-reduction distillation,
guidance distillation, and structured pruning),
parameter compression and branch-level calibration are addressed separately or
not at all; among these, \dash alone
constrains the guidance branches while reducing the parameter count.

\section{Method}
\label{sec:method}

Compressing a guided diffusion model exposes a property of the
distillation objective itself: supervising only the composite guided
score leaves the branch decomposition \textit{non-identifiable}
(\S\ref{sec:distil}). \dash resolves this through dual-branch
supervision, which constrains the conditional and unconditional
predictions independently; a frozen per-timestep curriculum copied
from the teacher, TIRT Transfer, accompanies it as a training-recipe
component and is described in the same section.

\subsection{Diffusion Model Preliminaries}
\label{sec:prelim}

Diffusion models~\cite{ddpm} corrupt data as
$\xt=\sqrt{\bar\alpha_t}\xo+\sqrt{1-\bar\alpha_t}\eps$ under a cosine
schedule~\cite{iddpm} with $T{=}1000$ steps, and train a network
$\eps_\theta(\xt,t,y)$ to predict $\eps$ under a per-timestep weighting
$\omega_t$; sampling is deterministic via DDIM~\cite{ddim} in
$K{\ll}T$ steps.

Classifier-free guidance~\cite{cfg} extrapolates away from
unconditional predictions:
\begin{equation}
  \tilde\eps = \eps_\theta(\xt,t,\varnothing) +
  w\underbrace{(\eps_\theta(\xt,t,y)-\eps_\theta(\xt,t,\varnothing))}
  _{\Delta\;:\;\text{guidance gap}},
\label{eq:cfg}
\end{equation}
where $w{>}1$ amplifies class-conditional signal. The guided
prediction $\tilde\eps$ is used at every denoising step, so
miscalibration in $\Delta$ accumulates over the generation
trajectory and is amplified by larger guidance weights. The
signal-to-noise ratio $\snr_t{=}\bar\alpha_t/(1{-}\bar\alpha_t)$
spans $\snr_0{\approx}10^4$ to $\snr_T{\approx}0$, motivating
the SNR-based reweighting introduced in \S\ref{sec:teacher}.

\subsection{\texorpdfstring{\textbf{\textsc{DASH}}}{DASH} Framework Overview}
\label{sec:overview}

\begin{figure}[tbp]
\centering
\figorbox{0.95\textwidth}{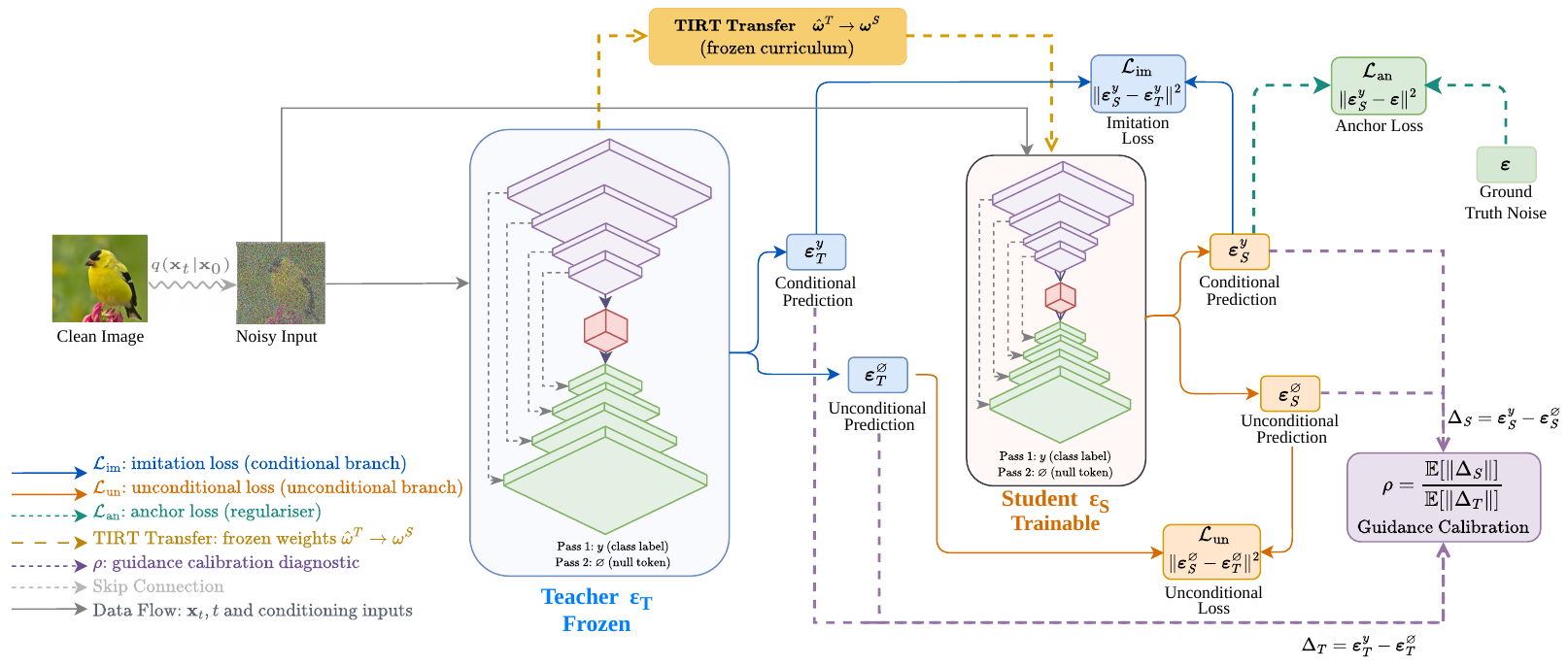}
\caption{\dash distillation pipeline. Frozen teacher (35.8M) and
trainable student (6.1M) each perform dual forward passes per sample.
\Lim and \Lun supervise the conditional and unconditional branches;
\Lan anchors the student to true noise $\eps$. TIRT Transfer (gold)
copies $\hat\omega^T$ as a frozen curriculum. At inference only the
student is retained.}
\label{fig:overview}
\end{figure}

Figure~\ref{fig:overview} depicts the training pipeline. The frozen
teacher $\eps_T$ and trainable student $\eps_S$ each execute two
forward passes per training sample: one with class label $y$ forced
(bypassing dropout) producing $\eps^y_T,\eps^y_S$, and one with null
token $\varnothing$ producing $\eps^\varnothing_T,\eps^\varnothing_S$.
Three losses supervise the student: \Lim for the conditional branch,
\Lun for the unconditional branch, and \Lan
as a data anchor. TIRT Transfer initialises the student's
per-timestep weights from the teacher's final learned values
$\hat\omega^T$, frozen throughout training. The guidance gap
ratio~$\rho$ (\S\ref{sec:impl}) monitors magnitude calibration
during training.

The student achieves $5.9{\times}$ compression (35.8M${\to}$6.1M)
through simultaneous channel reduction ($128{\to}64$) and depth
reduction ($\mathrm{nrb}\;2{\to}1$), following established practice
for UNet compression~\cite{bksdm,snapfusion}; self-attention at
$16{\times}16$ resolution is retained following the teacher's
configuration.

\subsection{Teacher: TIRT and TAG}
\label{sec:teacher}

The teacher employs an ADM~\cite{adm} UNet architecture with 128
base channels, 2~residual blocks per resolution ($\mathrm{nrb}{=}2$),
self-attention at $16{\times}16$ spatial resolution, totalling 35.8M
parameters.

\textbf{TIRT} extends Min-SNR~\cite{minsnr} with learnable
per-timestep scaling. Where Min-SNR's reweighting
$\min(\snr_t,\gamma)/\snr_t$ is fixed by the noise schedule alone and
therefore data-agnostic, TIRT learns $\lambda{\in}\mathbb{R}^T$
jointly with network parameters:
\begin{equation}
  \omega_t = \frac{\min(\snr_t,\gamma)}{\snr_t}\cdot\sigma(\lambda_t),
  \qquad \hat\omega_t = \frac{\omega_t}{\bar\omega},
\label{eq:tirt}
\end{equation}
where $\sigma(\cdot)$ is sigmoid, $\gamma{=}5.0$ caps the SNR
component, and $\bar\omega$ normalises to unit mean. Initialising
$\lambda_t{=}0$ gives $\sigma(0){=}0.5$ uniformly, which after
normalisation recovers the Min-SNR schedule up to a constant factor:
the curriculum starts there and redistributes emphasis from it, with
the sigmoid bounding $\omega_t$ against extreme reweighting. The TIRT objective
$\mathcal{L}_\mathrm{TIRT}{=}\mathbb{E}[\hat\omega_t\|\eps{-}
\eps_\theta\|^2]$ redistributes gradient signal across the
mid-trajectory timesteps, where the base schedule is flat and the
learned term dominates. The curriculum is learned end-to-end rather than
specified by hand.

\textbf{TAG} (Timestep-Adaptive Guidance) modulates guidance strength
at inference rather than training-time emphasis,
addressing the asymmetric requirements across the noise trajectory
identified in~\cite{cfgpp,intervalcfg}: low guidance at high noise
for structure formation ($t{\approx}T$), high guidance at low noise
for detail refinement ($t{\approx}0$). The teacher trains with
standard classifier-free guidance label dropout and applies a
sigmoid schedule over $w$ only during DDIM inference,
\begin{equation}
  w(t) = w_{\min} + (w_{\max}{-}w_{\min})\,
         \sigma\!\bigl(\kappa(0.5{-}t/T)\bigr),
\label{eq:tag}
\end{equation}
with $w_{\min}{=}1.0$ at high noise, $w_{\max}{=}4.0$ at low noise, and
steepness $\kappa{=}5.0$.
The branch-level distillation targets $\eps^y_T$ and
$\eps^\varnothing_T$ are produced by individual forward passes with
label $y$ and null token $\varnothing$ respectively. TAG is a shared
inference-time schedule applied to the teacher and all
two-branch students unless a fixed-$w$ result is explicitly labelled;
it does not change the branch targets used during distillation.

\subsection{Dual-Branch Output Distillation}
\label{sec:distil}

\subsubsection{The Null Space of Composite Distillation}

Conventional guided distillation~\cite{meng2023distill} supervises
the composite score
$\tilde\eps_S{=}\eps^\varnothing_S{+}w\Delta_S$ to match
$\tilde\eps_T$. This objective
$\mathcal{L}_{\mathrm{comp}}{=}\|\tilde\eps_S{-}\tilde\eps_T\|^2$
is non-identifiable at the branch level: any pair
$(\eps^y_S,\eps^\varnothing_S)$ satisfying
\begin{equation}
w\Delta_S = w\Delta_T + (\eps^\varnothing_T - \eps^\varnothing_S)
\label{eq:underdet}
\end{equation}
achieves zero loss. Writing the per-branch residuals as
$\delta^y{=}\eps^y_S{-}\eps^y_T$ and
$\delta^\varnothing{=}\eps^\varnothing_S{-}\eps^\varnothing_T$,
\eqref{eq:underdet} is equivalent to
$w\delta^y{+}(1{-}w)\delta^\varnothing{=}0$, a $d$-dimensional
linear subspace: every zero-loss pair drifts in the fixed ratio
$\delta^y/\delta^\varnothing{=}(w{-}1)/w$, so the set is the graph of
a single linear map. It contains both ideal recovery
(\textit{i}, $\delta^\varnothing{=}0$) and guidance collapse
(\textit{iii}, $\delta^\varnothing{=}w\Delta_T$), where
$\eps^y_S{\approx}\eps^\varnothing_S$ and CFG is inert at
inference. One linear constraint on two branches leaves
$\eps^\varnothing_S$ without a unique target. The ambiguity is not
confined to the zero-loss set: at each fixed input the loss is
\emph{exactly invariant} along this branch-output direction, at every
residual value (Proposition~C.1), so composite
supervision never distinguishes the ideal solution from collapse and
the argument does not require the student to reach zero loss. The
statement concerns the emitted branches in output space, not the
shared network parameters. Which point a composite student reaches is measured directly in
\S\ref{sec:distill_results}: its residuals sit on this line,
a third to two-fifths of the way toward collapse.

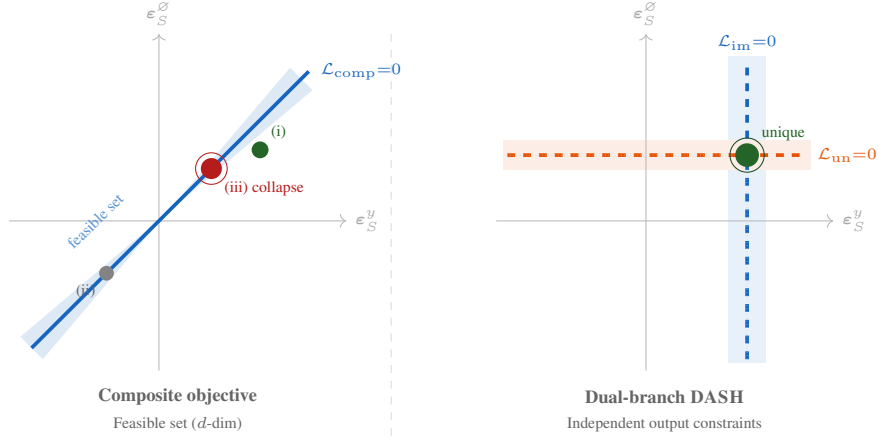
\begin{figure}[tbp]
\centering
\scalebox{0.99}{%
\begin{tikzpicture}[font=\small]
\begin{scope}[xshift=0mm]
  \fill[tblue!14](-1.55,-1.85)--(-1.85,-1.55)--(2.05,1.75)--(1.75,2.05)--cycle;
  \draw[->,gray!55,thin](-2.0,0)--(2.5,0) node[right,font=\tiny,gray!70]{$\eps^y_S$};
  \draw[->,gray!55,thin](0,-2.0)--(0,2.5) node[above,font=\tiny,gray!70]{$\eps^\varnothing_S$};
  \draw[tblue,line width=1.4pt](-1.7,-1.7)--(2.0,2.0);
  \node[font=\tiny,tblue!70,rotate=44] at (-0.85,0.0){feasible set};
  \fill[tgray!80](-0.7,-0.7)circle(2.8pt);
  \node[font=\tiny,tgray,below left=1pt] at (-0.7,-0.7){(ii)};
  \fill[tred](0.7,0.7)circle(4pt);
  \draw[tred,thin](0.7,0.7)circle(6pt);
  \node[font=\tiny,tred,align=left,below right=2pt] at (0.7,0.7){(iii) collapse};
  \fill[tgreen!80!black](1.35,0.95)circle(3.2pt);
  \node[font=\tiny,tgreen!80!black,above right=1pt] at (1.35,0.95){(i)};
  \node[font=\tiny,tblue,right=1pt] at (2.0,2.0){$\mathcal{L}_{\mathrm{comp}}{=}0$};
  \node[font=\scriptsize\bfseries,tgray] at (0.25,-2.35){Composite objective};
  \node[font=\tiny,tgray] at (0.25,-2.72){Feasible set ($d$-dim)};
\end{scope}
\begin{scope}[xshift=65mm]
  \fill[tblue!10](1.1,-1.9)rectangle(1.6,2.2);
  \fill[torange!10](-1.9,0.68)rectangle(2.2,1.08);
  \draw[->,gray!55,thin](-2.0,0)--(2.5,0) node[right,font=\tiny,gray!70]{$\eps^y_S$};
  \draw[->,gray!55,thin](0,-2.0)--(0,2.5) node[above,font=\tiny,gray!70]{$\eps^\varnothing_S$};
  \draw[tblue,line width=1.4pt,dashed](1.35,-1.85)--(1.35,2.15) node[above,font=\tiny,tblue]{$\Lim{=}0$};
  \draw[torange,line width=1.4pt,dashed](-1.85,0.88)--(2.15,0.88) node[right,font=\tiny,torange]{$\Lun{=}0$};
  \draw[gray!35,very thin,densely dotted](1.35,-1.85)--(1.35,0.88);
  \draw[gray!35,very thin,densely dotted](-1.85,0.88)--(1.35,0.88);
  \fill[tgreen!80!black](1.35,0.88)circle(4.5pt);
  \draw[tgreen!60!black,thin](1.35,0.88)circle(6.5pt);
  \node[font=\tiny,tgreen!80!black,above right=3pt] at (1.35,0.88){unique};
  \node[font=\scriptsize\bfseries,tgray] at (0.25,-2.35){Dual-branch \dash};
  \node[font=\tiny,tgray] at (0.25,-2.72){Independent output constraints};
\end{scope}
\draw[gray!28,thin,dashed](31mm,2.5)--(31mm,-2.9);
\end{tikzpicture}}
\caption{Solution space geometry. \textbf{Left}: Composite loss
admits a $d$-dimensional family of branch decompositions: the ideal
solution (\textit{i}, green), generic shifts (\textit{ii}, grey) and
collapse (\textit{iii}, red) are all points of the same feasible
set.
\textbf{Right}: Dual-branch constraints $\Lim$ and $\Lun$ specify
unique target outputs per sample, forcing $\Delta_S{\approx}\Delta_T$.}
\label{fig:underdet}
\end{figure}

This is a property of the objective
rather than of any particular architecture: whenever $\Delta_T{\neq}0$ and
no additional constraint acts on the unconditional branch,
supervising only composite or conditional scores each admits a
distinct $d$-dimensional ambiguity set, both containing collapse. A single fixed guidance weight is the source
of the non-identifiability: supervising at two distinct weights
resolves it algebraically, but conditioning degrades as those weights
converge, and \dash is the best-conditioned member of that family
(Corollary~C.3 and Remark~C.4).
Figure~\ref{fig:underdet} (left) illustrates the feasible solution
set.

\subsubsection{Solution: Independent Branch Supervision}

\dash resolves this non-identifiability by supervising both branches
independently. For training sample $(\xo,y,t)$ with corrupted
input $\xt$ and ground-truth noise $\eps$, the frozen teacher
provides $\eps^y_T{=}\eps_T(\xt,t,y)$ and
$\eps^\varnothing_T{=}\eps_T(\xt,t,\varnothing)$ with gradients
disabled; the student generates $\eps^y_S{=}\eps_S(\xt,t,y)$
with forced label and
$\eps^\varnothing_S{=}\eps_S(\xt,t,\varnothing)$
with null token. Three losses weighted by the frozen TIRT curriculum
$\hat\omega_t$ govern training:
\begin{align}
  \Lim  &= \mathbb{E}\bigl[\hat\omega_t
    \|\eps^y_S - \eps^y_T\|^2\bigr],             \label{eq:lim}\\
  \Lun  &= \mathbb{E}\bigl[\hat\omega_t
    \|\eps^\varnothing_S - \eps^\varnothing_T\|^2\bigr], \label{eq:lun}\\
  \Lan  &= \mathbb{E}\bigl[\hat\omega_t
    \|\eps^y_S - \eps\|^2\bigr],                  \label{eq:lan}\\
  \Ltot &= \Lim + \Lun + \lambda_\mathrm{an}\Lan,  \label{eq:total}
\end{align}
where $\lambda_\mathrm{an}{=}0.1$ and $\eps$ denotes the
ground-truth noise sample used to corrupt $\xo$, not any
teacher prediction.

\Lim requires $\eps^y_S{=}\eps^y_T$; \Lun requires
$\eps^\varnothing_S{=}\eps^\varnothing_T$. For a fixed input
$(\xt,t,y)$, these constraints uniquely specify the target outputs
for each branch, reducing the feasible set
of~\eqref{eq:underdet} from a $d$-dimensional subspace to a single
point. When both $\Lim{=}0$ and $\Lun{=}0$, the
student must satisfy $\eps^y_S{=}\eps^y_T$ \textit{and}
$\eps^\varnothing_S{=}\eps^\varnothing_T$, excluding shifts and
collapse as feasible solutions when $\Delta_T{\ne}0$
(Figure~\ref{fig:underdet}, right; proved in
\S\ref{app:proof_underdet}).

\Lan anchors the conditional branch to ground-truth noise,
regularising against teacher error on the branch that carries the
class signal. Setting $\lambda_\mathrm{an}{=}0.1$ keeps
\Lan a corrective prior without dominating distillation.
Branch loss coefficients are symmetric
($\lambda_\mathrm{im}{=}\lambda_\mathrm{un}{=}1.0$). This is not an
arbitrary choice: at the output-residual level it makes their Gram
matrix the identity, avoiding an explicit branch-weight imbalance; the anchor adds a third,
conditional-only term outside this pair. Asymmetric weights leave both branches
identifiable but unequally weighted in the gradient, the same
conditioning penalty that separates composite variants
(\S\ref{sec:distill_results}). \S\ref{sec:experiments} measures the recovered gap
directly and isolates \Lun's contribution through ablation.

\textbf{TIRT Transfer.}
Per-timestep weights $\lambda{\in}\mathbb{R}^T$ need hundreds of
epochs to converge, co-adapting with network parameters. Because teacher training and
student distillation induce different per-timestep error profiles,
their schedule updates need not agree (\S C.2). The teacher's
final normalised schedule is therefore transferred directly,
$\omega^S_t\leftarrow\hat\omega^T_t$ for all $t$, and held fixed
(excluded from the optimiser). Both halves are measured within the
tested budget: relearning the schedule from $\lambda^S{=}0$ costs
$3.84$\,/\,$3.94$ FID at 234K iterations (Table~\ref{tab:ablation}),
and the schedule itself earns the teacher $2.05$\,/\,$2.47$ FID over
fixed Min-SNR weighting at matched architecture, data and budget
(Table~D8).


\section{Experiments}
\label{sec:experiments}

The experiments show that explicit unconditional supervision
preserves guidance calibration under parameter compression where
single-constraint objectives~\cite{meng2023distill} and
alternative compression strategies fail.

\subsection{Setup}
\label{sec:impl}

CIFAR-10 and CIFAR-100~\cite{cifar10} provide $32{\times}32$ RGB
images over 10 and 100 classes respectively, giving complementary
evaluation at two levels of class granularity.
ImageNet-64~\cite{imagenet64} extends the
evaluation to 1000 classes at $64{\times}64$ resolution
(1{,}281{,}167 training images), a substantially larger label space
and higher resolution than either CIFAR setting.

The primary student (64 base channels, 6.1M parameters,
$5.9{\times}$ compression) distils from the frozen teacher EMA, with
48- and 96-channel widths additionally evaluated to assess compression
scaling; iteration counts, loss weights, and optimiser settings are
tabulated in \S\ref{app:impl}.
Inference is profiled on a single NVIDIA T4: the
student reduces FLOPs by $5.77{\times}$ and increases throughput by
$3.62{\times}$ at batch size 64.

Six supervision baselines share the student architecture, frozen
teacher, TIRT schedule, optimiser settings and iteration budget; they
differ in the supervision signal and, for label-dropout, in the number
of teacher branch queries per step (Table~\ref{tab:distill}). \textit{Scratch} is a separate teacher-free reference
trained under the fixed Min-SNR schedule, iteration-matched and
carrying no teacher cost.
\textit{Composite} matches the full guided score at $w{=}4.0$, a
parameter-compression adaptation of guided-output
distillation~\cite{meng2023distill}; \textit{multi-$w$ composite}
samples $w$ per step from $[1.5,5.0]$ for the same target;
\textit{conditional-only} supervises the conditional branch alone,
leaving the unconditional branch unsupervised; \textit{gap matching}
constrains $\Delta_S{-}\Delta_T$ without anchoring either branch.
Each loss is written out in Table~A6.
\textit{Label-dropout} matches one teacher branch per example, selected
by the teacher's own null-token dropout, and so issues half as many
teacher queries per step as \dash.
\textit{FitNets}~\cite{fitnets} adds intermediate UNet feature
matching to composite supervision.

Two further baselines vary the compression strategy rather than the
loss. Neither receives TIRT Transfer.
\textit{Magnitude pruning} applies structured channel pruning to the
trained teacher followed by finetuning under a matched budget, using
the standard magnitude criterion rather than Diff-Pruning's
Taylor-based objective~\cite{diffpruning}.
\textit{Single-head, $w$-conditioned} distils the teacher's composite guided
output into one network conditioned directly on $w$, following
Meng~\etal~\cite{meng2023distill}; it discards the branch
decomposition entirely and therefore exposes no separate conditional
or unconditional prediction at inference.

Unless a table states otherwise, all results use 50K samples at
50-step DDIM with TAG $w_{\max}{=}4.0$ and 10K matched calibration
pairs. Tables~\ref{tab:distill}--\ref{tab:ablation} report
FID~\cite{fid}, IS~\cite{is} and the calibration metrics as means over
three independent student runs (seeds 42\,/\,123\,/\,456), all
distilled from the same frozen teacher checkpoint. Three quantities summarise how faithfully
the student reproduces the teacher's guidance gap: the magnitude
ratio $\rho{=}\mathbb{E}[\|\Delta_S\|]/\mathbb{E}[\|\Delta_T\|]$,
the directional cosine $\cos(\Delta){=}\mathbb{E}[\Delta_S^{\!\top}
\Delta_T/(\|\Delta_S\|\|\Delta_T\|)]$, and the absolute error
$\mathrm{MSE}_\Delta{=}\mathbb{E}[\tfrac{1}{d}\|\Delta_S-\Delta_T\|^2]$
over the $d$ output elements. Both $\rho$ and
$\cos(\Delta)$ are required: a high $\rho$ alone does not preclude a
misaligned gap (\textit{Scratch}, $\rho{=}0.80$ with
$\cos(\Delta){=}0.51$).
The single-head baseline emits no separate branch predictions, so all
three are undefined for it.

\subsection{Distillation Results}
\label{sec:distill_results}

Table~\ref{tab:distill} presents the core comparison. The teacher
reaches FID~5.47\,/\,6.80 (IS~9.42\,/\,7.46) on CIFAR-10/100 at 50 steps and
35.8M parameters; published models reach lower absolute FID under different
architectures, training recipes and sampling budgets
(Table~D8), so the teacher
trades absolute quality for a fixed, moderate budget the student
inherits; all comparisons here are relative under that budget.

Scratch and composite distillation bound the FID range: distillation
helps, but composite's guidance gap survives only partially, since
the composite objective is satisfied by branch decompositions where
$\Delta_S{\neq}\Delta_T$~\eqref{eq:underdet}. Its
$\mathrm{MSE}_\Delta$ is
several times higher than \dash's, consistent with collapse
following from the objective rather than from an incidental
training artefact.
Conditional-only supervision is the sharpest case: full teacher
supervision on the conditional branch alone still collapses the
gap, falling below scratch on both datasets. The null-token embedding
receives no gradient under $\mathcal{L}_\mathrm{cond}$, so
$\eps^\varnothing_S$ is shaped only by the shared trunk; the branches
converge and $\rho$ falls to $0.09$. At TAG $w_{\max}{=}4.0$ this
amplifies a small, misaligned $\Delta_S$ at every denoising step,
so guidance perturbs the sample rather than sharpening it.

Varying the guidance weight during training recovers part of the gap.
Multi-$w$
composite lifts $\rho$ from composite's
$0.68$\,/\,$0.65$ to $0.75$\,/\,$0.71$ and cuts $\mathrm{MSE}_\Delta$
by $18\%$ on CIFAR-10, where the FID difference is within training-seed
variation, yet stops well short of \dash's $0.91$\,/\,$0.89$.
Sampling $w$ makes the objective identifiable but strongly
ill-conditioned (condition number ${\approx}304$ against $1$ for
symmetric dual-branch supervision, Remark~C.4).
Distilling under the teacher's own null-token dropout stays in the
well-conditioned basis but observes one branch per example, so both
constraints hold only in expectation. It reaches
$\rho{=}0.81$\,/\,$0.82$, and on CIFAR-100 its $\cos(\Delta)$ falls
below its own $\rho$: a branch seen on a tenth of samples fixes the
gap's scale before its direction.
The penalty for replacing TAG with a fixed guidance weight scales with
how poorly the gap is calibrated, from $0.63$ FID for the teacher to
$1.48$ without \Lun (Table~D16).

The residuals confirm that composite students land where the null
space predicts. \eqref{eq:underdet} forces the branch-residual ratio
to $(w{-}1)/w{=}0.75$ at $w{=}4$ whatever the drift, and a solution a
fraction $f$ toward collapse has $\rho{=}1{-}f$ on the line, hence
$f{\geq}1{-}\rho$ in general. Measured on the three datasets the ratio
is $0.77$, $0.73$ and $0.79$, and $f$ is $0.34$, $0.38$ and $0.40$
against $1{-}\rho$ of $0.32$, $0.35$ and $0.35$: both hold, with the
excess from error orthogonal to the line (\S C.1).

Constraining elsewhere does not close the gap either. FitNets keeps composite supervision and adds intermediate
feature matching on top of it, yet calibrates worse than composite
alone: aligning internal representations places no constraint on the
emitted branches, which remain as under-determined as before. With the
feature term removed it reduces to composite supervision, reaching
$\rho{=}0.68$; adding the term lowers this to $0.63$.
Gap matching constrains only the difference: its zero-loss set is
$\delta^y{=}\delta^\varnothing$, so both branches may drift together.
It optimises $\mathrm{MSE}_\Delta$ directly yet reaches $0.063$ where
\dash, which never targets that quantity, reaches $0.028$: constraining
each branch is a better route to a calibrated gap than constraining the
gap, because the difference objective carries no anchor for either
branch. Its margin to \dash widens on CIFAR-100.

Among the configurations compared, \dash alone holds both axes at once:
class-conditional separation is preserved and the guidance direction
stays aligned with the teacher's. The residual undercalibration in
$\rho$ shows up as texture softening rather than class confusion
(\S\ref{sec:qualitative}), and the CIFAR-100 margin over composite
distillation ($10.47$ against $20.84$) is wider than the CIFAR-10
margin ($8.87$ against $13.84$).

\textbf{ImageNet-64.}
Table~\ref{tab:in64} repeats the core comparison at 1000 classes.
Composite distillation reaches $\rho{=}0.65$ and FID~18.43 against
\dash's $0.89$ and $9.10$: the same $0.24$ deficit seen at 100 classes. Removing \Lun alone costs
11.03 FID here. Neither the non-identifiability nor its correction is specific to
$32{\times}32$ inputs.

\textbf{Alternative compression strategies.}
Magnitude pruning inherits teacher weights directly, at $47\%$ more
parameters than the \dash student and under the same finetuning
budget. Inheriting the teacher's branch structure improves on composite in
magnitude ($\rho$ $0.71$ versus $0.68$) but not in direction
($\cos(\Delta)$ $0.68$ versus $0.71$), and produces worse samples on FID: inherited weights are not a
substitute for constraining the branches during training.

The single-head baseline collapses the branches into one
$w$-conditioned head, following the guidance-distillation stage of
Meng~\etal~\cite{meng2023distill}, and tests whether the branch
decomposition is worth keeping under compression at all. \dash improves on it by $7.91$ FID on CIFAR-10 and $13.03$ on
CIFAR-100 while using $32\%$ fewer parameters. The
trade is not one-sided: \dash is the smaller model, but single-head
runs one forward pass per step where \dash runs two, so its per-step
FLOPs are lower (Table~A3); the measured wall-clock margin is
comparable to profiling variance. It also exposes no branches for guidance operators at inference.

\begin{table}[tbp]
\caption{Main distillation results, FID$\downarrow$\,/\,IS$\uparrow$. All 6.1M rows
except \textit{Scratch} share training conditions (\S\ref{sec:impl}); compression-strategy
baselines differ in size.
$^\ast$$\cos(\Delta)$ not meaningful when
$\|\Delta_S\|{\approx}0$~($\rho{<}0.10$); $^\dagger$single-head emits no
separate branches, so all three are undefined for it. $\rho$ is a ratio of expectations and $\cos(\Delta)$ a per-sample mean,
while $\mathrm{MSE}_\Delta$ is a second moment; the three are not
algebraically determined by one another in general (\S D.4).
$^\ddagger$$w$ sampled per step rather than fixed
(\S\ref{sec:impl}).}
\label{tab:distill}
\centering\small\setlength{\tabcolsep}{4pt}
\resizebox{\textwidth}{!}{%
\begin{tabular}{lcccccccc}
\toprule
& & \multicolumn{2}{c}{CIFAR-10} & \multicolumn{2}{c}{CIFAR-100} &
  \multicolumn{3}{c}{Calibration (C10\,/\,C100)} \\
\cmidrule(lr){3-4}\cmidrule(lr){5-6}\cmidrule(lr){7-9}
Model & Params & FID$\downarrow$ & IS$\uparrow$ &
  FID$\downarrow$ & IS$\uparrow$ &
  $\rho$ & $\cos(\Delta)$ & $\mathrm{MSE}_\Delta$ \\
\midrule
\multicolumn{9}{l}{\textit{Teacher:}} \\
\dash-Teacher & 35.8M
  & 5.47 & 9.42 & 6.80 & 7.46
  & 1.00 & 1.00 & 0.000 \\
\midrule
\multicolumn{9}{l}{\textit{Loss-formulation baselines
  (6.1M, $5.9{\times}$):}} \\
Scratch
  & 6.1M & 20.47 & 7.44 & 26.14 & 6.51
  & 0.80\,/\,0.78 & 0.51\,/\,0.48 & 0.213\,/\,0.247 \\
Cond-only $\eps^y$
  & 6.1M & 22.31 & 7.21 & 27.83 & 5.98
  & 0.09\,/\,0.08 & ---$^\ast$
  & 0.242\,/\,0.292 \\
Composite $\tilde\eps$~\cite{meng2023distill}
  & 6.1M & 13.84 & 8.63 & 20.84 & 6.84
  & 0.68\,/\,0.65 & 0.71\,/\,0.69 & 0.119\,/\,0.152 \\
Multi-$w$ composite$^\ddagger$
  & 6.1M & 13.67 & 8.96 & 19.13 & 7.23
  & 0.75\,/\,0.71 & 0.79\,/\,0.77
  & 0.098\,/\,0.140 \\
FitNets~\cite{fitnets}
  & 6.1M & 12.84 & 8.74 & 19.47 & 6.89
  & 0.63\,/\,0.61 & 0.68\,/\,0.66
  & 0.128\,/\,0.161 \\
Gap $\Delta$ match
  & 6.1M & 11.42 & 8.34 & 18.46 & 6.58
  & 0.81\,/\,0.78 & 0.86\,/\,0.84
  & 0.063\,/\,0.086 \\
Label-dropout
  & 6.1M & 11.43 & 8.88 & 14.56 & 7.10
  & 0.81\,/\,0.82 & 0.83\,/\,0.79
  & 0.059\,/\,0.071 \\
\midrule
\multicolumn{9}{l}{\textit{Alternative compression strategies:}} \\
Magnitude pruning
  & 8.95M & 18.04 & 8.56 & 24.48 & 6.89
  & 0.71\,/\,0.66 & 0.68\,/\,0.65
  & 0.080\,/\,0.101 \\
Single-head, $w$-conditioned
  & 8.99M & 16.78 & 9.00 & 23.50 & 7.19
  & ---$^\dagger$ & ---$^\dagger$
  & ---$^\dagger$ \\
\midrule
\dash~(ours)
  & 6.1M
  & \textbf{8.87} & \textbf{9.31}
  & \textbf{10.47} & \textbf{7.41}
  & 0.91\,/\,0.89 & 0.94\,/\,0.93 & 0.028\,/\,0.040 \\
\bottomrule
\multicolumn{9}{l}{%
  \footnotesize At $5.9{\times}$ compression \dash gains
  11.60\,/\,15.67 FID over scratch and 4.97\,/\,10.37 over composite.}
\end{tabular}}
\end{table}

\begin{table}[tbp]
\caption{Class-conditional ImageNet-64 (1000 classes,
$64{\times}64$); 116.5M teacher, 22.3M student ($5.2{\times}$, distinct
from the $5.9{\times}$ CIFAR setting). Means, with std in
Table~D10. $^\S$Isolates \Lun as the source of the improvement.}
\label{tab:in64}
\centering\small\setlength{\tabcolsep}{4pt}
\begin{tabular}{lccccc}
\toprule
Model & FID$\downarrow$ & IS$\uparrow$ &
  $\rho$ & $\cos(\Delta)$ & $\mathrm{MSE}_\Delta$ \\
\midrule
Teacher & 6.34 & 48.35 & 1.00 & 1.00 & 0.000 \\
\midrule
Composite~\cite{meng2023distill} & 18.43 & 43.67 &
  0.65 & 0.68 & 0.130 \\
w/o \Lun$^\S$ & 20.13 & 41.67 &
  0.22 & 0.27 & 0.228 \\
\midrule
\dash~(ours) & \textbf{9.10} & \textbf{47.32} &
  \textbf{0.89} & \textbf{0.92} & \textbf{0.036} \\
\bottomrule
\end{tabular}
\end{table}

\subsection{Ablation Study}
\label{sec:ablation}

Table~\ref{tab:ablation} isolates each component. The distinguishing
component is \Lun: its removal costs 9.97\,/\,12.48 FID and collapses
$\rho$ to 0.23\,/\,0.24 on CIFAR-10 and CIFAR-100 (the anchor's larger
effect under collapse is analysed in \S D.3).
Although removing \Lim yields the largest absolute FID degradation,
\Lim is the conventional imitation objective shared by all
output-level baselines; \Lun is what distinguishes \dash from them.
The IS drop from 9.31 to 7.23 on CIFAR-10 indicates
that
without explicit $\boldsymbol{\varepsilon}^\varnothing$ matching,
the conditional--unconditional separation required for effective CFG
degrades independently of conditional branch quality.
Without \Lim, $\boldsymbol{\varepsilon}^y_S$ is anchored only by
\Lan, toward ground-truth noise rather than the teacher: even with
correct unconditional predictions from \Lun, the gap points in the
wrong direction ($\cos(\Delta){=}0.46$\,/\,0.43,
$\rho{=}0.41$\,/\,0.38), reaching FID~21.84\,/\,28.37, below
scratch on both datasets.
The anchor \Lan contributes 1.67\,/\,1.91 FID, secondary to branch
supervision.

The TIRT Transfer rows decompose the curriculum contribution.
Replacing frozen teacher curriculum with frozen Min-SNR weights
costs 2.60\,/\,2.67 FID and drops $\rho$ from 0.91 to 0.84,
demonstrating that the final learned curriculum carries information
beyond what the noise schedule alone specifies. Allowing
$\lambda_t$ to train from zero under \Ltot costs a further
1.24\,/\,1.27 FID on CIFAR-10 and CIFAR-100. \S C.2 gives the
schedule-gradient form and why the teacher's and the student's
per-timestep error profiles need not drive $\lambda_t$ to the same
place.

Repeating the comparison under a standard Min-SNR teacher widens the
margin over composite rather than narrowing it (9.58 versus 4.97 FID
points; \S D.6), so the non-identifiability is a property of the
distillation objective, not of teacher construction.

\begin{table}[tbp]
\caption{Component ablation, 6.1M student. All distilled and ablation
rows share the student architecture, frozen teacher, and
234K-iteration budget; \textit{Scratch} is a teacher-free reference
with the same architecture and budget.}
\label{tab:ablation}
\centering\small\setlength{\tabcolsep}{4pt}
\resizebox{\textwidth}{!}{%
\begin{tabular}{lcccccccccc}
\toprule
& \multicolumn{3}{c}{Components} &
  \multicolumn{2}{c}{CIFAR-10} &
  \multicolumn{2}{c}{CIFAR-100} &
  \multicolumn{3}{c}{Calibration (C10\,/\,C100)} \\
\cmidrule(lr){2-4}\cmidrule(lr){5-6}\cmidrule(lr){7-8}\cmidrule(lr){9-11}
Configuration & \Lim & \Lun & \Lan &
  FID$\downarrow$ & IS$\uparrow$ & FID$\downarrow$ & IS$\uparrow$ &
  $\rho$ & $\cos(\Delta)$ & $\mathrm{MSE}_\Delta$ \\
\midrule
Scratch
  & -- & -- & --
  & 20.47 & 7.44 & 26.14 & 6.51
  & 0.80\,/\,0.78 & 0.51\,/\,0.48 & 0.213\,/\,0.247 \\
w/o \Lun
  & \cmark & \xmark & \cmark
  & 18.84 & 7.23 & 22.95 & 6.21
  & 0.23\,/\,0.24 & 0.27\,/\,0.25
  & 0.221\,/\,0.271 \\
w/o \Lim
  & \xmark & \cmark & \cmark
  & 21.84 & 7.12 & 28.37 & 5.84
  & 0.41\,/\,0.38 & 0.46\,/\,0.43
  & 0.280\,/\,0.332 \\
w/o \Lan
  & \cmark & \cmark & \xmark
  & 10.54 & 9.11 & 12.38 & 7.28
  & 0.88\,/\,0.86 & 0.93\,/\,0.92
  & 0.033\,/\,0.046 \\
TIRT: fixed schedule
  & \cmark & \cmark & \cmark
  & 11.47 & 8.62 & 13.14 & 7.03
  & 0.84\,/\,0.81 & 0.89\,/\,0.87
  & 0.050\,/\,0.072 \\
TIRT: learned $\lambda_0{=}0$
  & \cmark & \cmark & \cmark
  & 12.71 & 8.44 & 14.41 & 6.88
  & 0.82\,/\,0.79 & 0.87\,/\,0.84
  & 0.059\,/\,0.082 \\
\dash~(ours)
  & \cmark & \cmark & \cmark
  & \textbf{8.87} & \textbf{9.31}
  & \textbf{10.47} & \textbf{7.41}
  & 0.91\,/\,0.89 & 0.94\,/\,0.93
  & 0.028\,/\,0.040 \\
\bottomrule
\end{tabular}}
\end{table}

FID and $\rho$ do not order the methods identically, in both
directions: FitNets beats Composite on FID (12.84 vs.\ 13.84) with
worse $\rho$ (0.63 vs.\ 0.68), while the teacher-free \textit{Scratch}
reference reaches $\rho{=}0.80$ on a gap that is largely intact but
misdirected ($\cos(\Delta){=}0.51$ against Composite's 0.71).
Composite, multi-$w$, FitNets and pruning all sit below it: neither
composite supervision nor weight inheritance holds the gap magnitude
above a model trained with no teacher at all. The deficit is therefore
not a consequence of reduced capacity, and no inference-time rescaling repairs
it: a scalar corrects a global magnitude mismatch but cannot recover
the teacher's guidance direction (Remark~C.5). \dash holds $\rho(t){\in}[0.80,\,0.96]$ across the trajectory
while composite collapses near-uniformly (Figure~D3a), and the
gains persist from $2.5{\times}$ to $10.5{\times}$ compression
(Table~D14).
\begin{figure}[tbp]
\centering
\setlength{\tabcolsep}{4pt}
\resizebox{\textwidth}{!}{%
\begin{tabular}{@{}cc@{}}
\figorbox{0.46\textwidth}{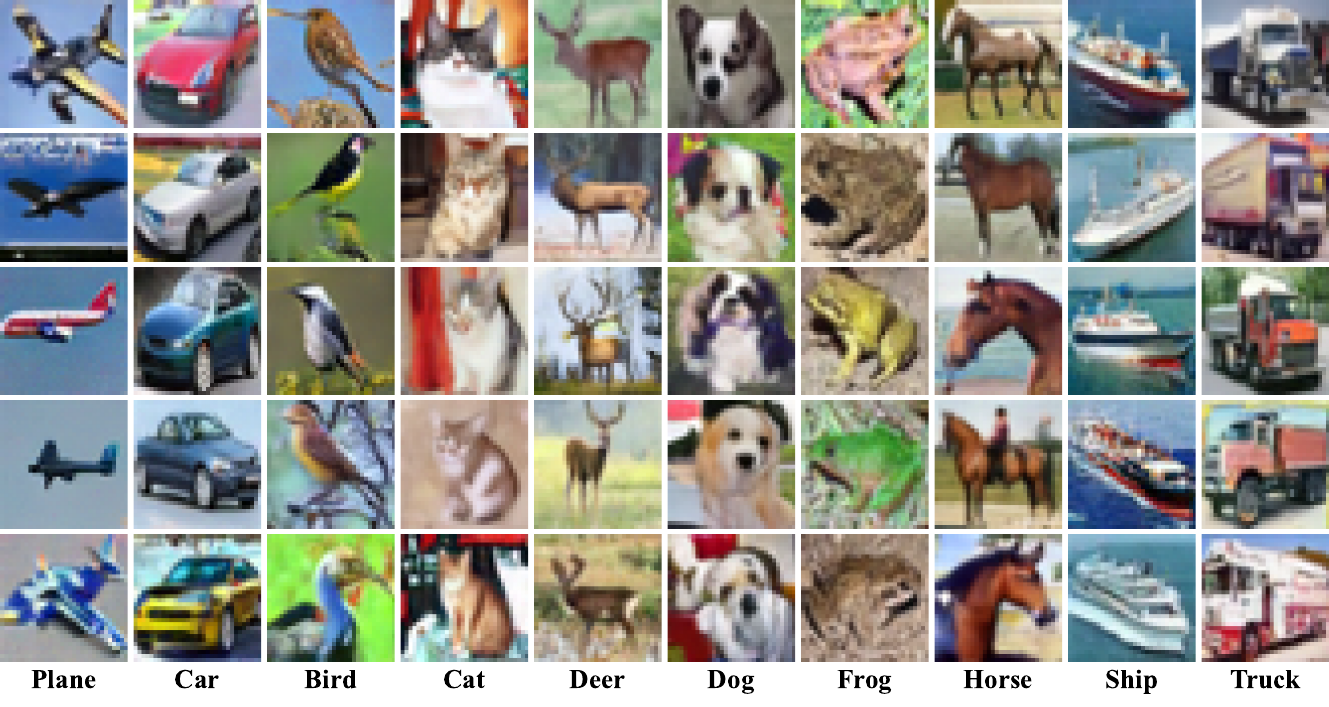} &
\figorbox{0.46\textwidth}{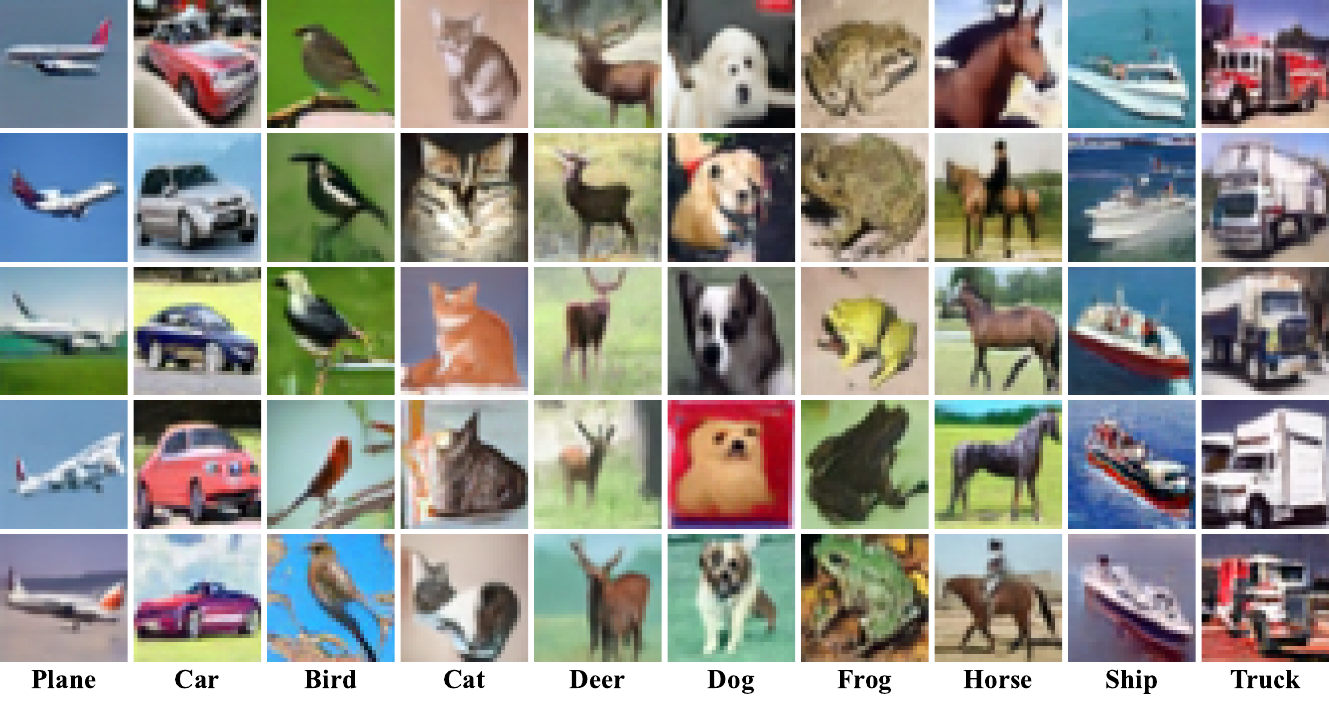} \\[1mm]
{\small (a)~Teacher (35.8M, FID~5.47)} &
{\small (b)~\dash student (6.1M, FID~8.87)}
\end{tabular}}
\caption{Teacher vs.\ \dash student, CIFAR-10 ($5{\times}10$ grids,
matched seeds, TAG $w_{\max}{=}4.0$, 50-step DDIM). Columns left to right:
Plane, Car, Bird, Cat, Deer, Dog, Frog, Horse, Ship, Truck.}
\label{fig:teacher_student}
\end{figure}

Because label-dropout issues one teacher query per step to \dash's
two, \dash was also trained for $117$K iterations, equalising the
teacher-query budget; it reaches FID $9.76$ and $\rho{=}0.89$ on
CIFAR-10 (single seed), ahead of label-dropout at the full budget on
both, and conservatively so, since halving iterations also halves the
optimiser steps. 
The same run shows calibration settling before sample
quality: $\rho$ falls only from $0.91$ to $0.89$ while FID degrades by
$0.89$ points, so the diagnostic reaches its final value while FID
improves.

\subsection{Qualitative Results}
\label{sec:qualitative}

The calibration diagnostic, measured on forward-noised inputs,
indicates where compression should cost quality: at $\rho{=}0.91$ the
student's guidance magnitude is $9\%$ below the teacher's, and since
guidance sharpens detail mainly at low-noise timesteps, texture should
soften while class structure holds. Figure~\ref{fig:teacher_student} confirms this:
deviation is concentrated mainly in fine surface detail, and a minority of
fine-grained bird and cat samples are ambiguous at $32{\times}32$ even
for the teacher.

\section{Conclusion}

Guidance collapse is a zero-loss solution of the standard composite
guided-score objective, and in output space that objective cannot
separate it from ideal recovery at any residual value. The failure
therefore follows from the supervision signal rather than from
initialisation, and composite students land on the predicted null line
at $5.9{\times}$ compression, so finite capacity does not absorb it
either.

Removing \Lun costs several times more than removing the anchor or the
curriculum transfer; \Lim costs more still, but is the imitation term
every output-level baseline already has
(Table~\ref{tab:ablation}). FitNets ($\rho{=}0.63$) shows separately
that internal feature supervision added to composite supervision is
still no substitute for a branch-level constraint. The TIRT Transfer ablation shows the teacher's curriculum is not
recovered within the distillation budget.

\textbf{Limitations.}
The ImageNet-64 evaluation (Table~\ref{tab:in64}) establishes that
the branch non-identifiability and its correction persist at
1000 classes and $64{\times}64$ resolution; the full baseline sweep and
ablations are conducted at CIFAR scale. Cross-attention conditioning in latent
diffusion~\cite{ldm} and transformer
backbones~\cite{peebles2023scalable} remain open. The
residual offset $\rho{=}0.91$ tracks capacity under
distillation: $\rho$ scales monotonically with student width
($0.86{\to}0.91{\to}0.94$) where the teacher-free reference stays flat
($0.81{\to}0.80{\to}0.83$, Table~D14).

The calibration pair $(\rho,\,\cos(\Delta))$ requires only paired
teacher and student outputs, so it applies to any guided distillation
pipeline, diagnosing branch collapse that FID alone does not
distinguish (\S\ref{sec:ablation}). Code is available at \url{https://github.com/C-loud-Nine/DASH_Dual-Branch-Score-Distillation}.

{
    \small

\bibliographystyle{iclr2026_conference}
\bibliography{main}

\newpage
\appendix
\setcounter{section}{0}
\renewcommand{\thesection}{\Alph{section}}
\renewcommand{\thetable}{\Alph{section}\arabic{table}}
\renewcommand{\thefigure}{\Alph{section}\arabic{figure}}
\setcounter{table}{0}
\setcounter{figure}{0}

\section*{Appendix Overview}
\label{app:roadmap}
The appendix is organised as follows.
\S\ref{app:impl} gives implementation and training details for
reproducibility. \S\ref{app:related} extends the related-work
positioning from the main text. \S\ref{app:theory} contains the full
proofs: branch non-identifiability and determinacy under two-weight
supervision in \S\ref{app:proof_underdet}, and the TIRT curriculum
argument in \S\ref{app:proof_tirt}. \S\ref{app:ablations} reports
extended ablations: teacher component
decomposition, per-seed standard deviations, per-timestep and
training-dynamics calibration views, step-count and guidance-scale
generalisation, width scaling, fixed-$w$ evaluation, and a repeat of
the core comparison under a standard Min-SNR teacher.
\S\ref{app:qualitative} provides additional qualitative samples:
class-conditional diversity grids, sampling-step comparisons, and
failure cases.
\S\ref{app:limitations} states limitations, broader impact, and the
LLM-usage statement.

\suppressfloats[t]

\section{Implementation Details}
\label{app:impl}

Tables~\ref{tab:arch_train} and~\ref{tab:eval_proto} list all
hyperparameters. Learning rate and gradient norm clipping at~1.0 follow the DDPM
protocol~\cite{ddpm}; AdamW ($\beta{=}(0.9,0.999)$) and cosine
annealing are ours.
\dash trains in 16h\,/\,17h on C10\,/\,C100. Label-dropout queries a
single teacher branch per step (Table~\ref{tab:repro_baselines}),
halving its teacher-side cost. Inference cost is unchanged for both.

\begin{table}[t]
\caption{Architecture and training hyperparameters. Paired values are
CIFAR-10\,/\,CIFAR-100. ($^{\S}$) \dash{} needs no dropout because both branches are
supervised on every step; the baselines that use it are listed in
Table~\ref{tab:repro_baselines}. The TIRT parameters $\lambda_t$ share
the network's AdamW group and learning rate.}
\label{tab:arch_train}
\centering\small\setlength{\tabcolsep}{4pt}
\begin{tabular}{lcc}
\toprule
Hyperparameter & Teacher & Student \\
\midrule
\multicolumn{3}{l}{\textit{Architecture}} \\
Base channels / mult.  & 128\,/\,$(1,2,2,2)$ & 64\,/\,$(1,2,2,2)$ \\
Res.\ blocks per res.  & 2 & 1 \\
Attention resolution   & $16{\times}16$ & $16{\times}16$ \\
Parameters             & 35.8M & 6.1M \\
\midrule
\multicolumn{3}{l}{\textit{Training}} \\
Learning rate     & $2{\times}10^{-4}$ & $1{\times}10^{-4}$ \\
Total iterations  & 391K (${\approx}500$ ep.) & 234K (${\approx}300$ ep.) \\
Batch / EMA $\tau$& 64\,/\,0.9999 & 64\,/\,0.9995 \\
CFG dropout       & 10\% & dual-pass (none)$^{\S}$ \\
Training time     & 24h\,/\,25h & 16h\,/\,17h \\
Hardware (train)  & \multicolumn{2}{c}{Single NVIDIA T4 16\,GB} \\
\midrule
\multicolumn{3}{l}{\textit{DASH loss weights (student only)}} \\
$\lambda_{\mathrm{im}}$\,/\,$\lambda_{\mathrm{un}}$ & -- & 1.0\,/\,1.0 \\
$\lambda_{\mathrm{an}}$ & -- & 0.1$^\dagger$ \\
\midrule
\multicolumn{3}{l}{\textit{TIRT}} \\
SNR cap $\gamma$  & 5.0 & frozen from teacher \\
$\lambda_t$ init. & learned & copied from teacher \\
\midrule
\multicolumn{3}{l}{\textit{TAG (inference)}} \\
$w_{\min}$\,/\,$w_{\max}$\,/\,$\kappa$ & \multicolumn{2}{c}{1.0\,/\,4.0\,/\,5.0} \\
\midrule
\multicolumn{3}{l}{\textit{ImageNet-64 (Table~\ref{tab:in64})}} \\
Base channels     & 128 & 64 \\
Res.\ blocks      & 3 & 2 \\
Channel mult.     & \multicolumn{2}{c}{$(1,2,3,4)$} \\
Attention res.    & \multicolumn{2}{c}{$16{\times}16$, $8{\times}8$} \\
Parameters        & 116.5M & 22.3M ($5.2{\times}$) \\
Epochs / batch    & 300\,/\,2048 & 130\,/\,2048 \\
LR schedule       & \multicolumn{2}{c}{cosine from $2{\times}10^{-4}$, no warmup} \\
Hardware          & \multicolumn{2}{c}{TPU v5e-8 (Table~\ref{tab:repro_env})} \\
\bottomrule
\multicolumn{3}{p{0.95\linewidth}}{\scriptsize $^\dagger$%
  $\lambda_{\mathrm{an}}$ selected from $\{0.01,0.05,0.1,0.2,0.5\}$;
  FID insensitive within $[0.05,0.2]$ (${\pm}0.3$ on C10).}
\end{tabular}
\end{table}

\noindent\textit{ImageNet-64 protocol.} ImageNet-64 uses the same loss
weights, TIRT settings, and evaluation protocol as the CIFAR runs; the
settings that differ are listed in the final block of the table above.
The student budget is identical for \dash and every ImageNet baseline,
so the comparison is budget-matched.

\begin{table}[t]
\caption{Shared evaluation protocol (teacher and student).}
\label{tab:eval_proto}
\centering\small\setlength{\tabcolsep}{3pt}
\begin{tabular}{ll}
\toprule
Setting & Value \\
\midrule
Sampler             & DDIM, $K{=}50$ steps, TAG $w_{\max}{=}4.0$ \\
FID\,/\,IS samples  & 50,000 \\
FID reference       & 50,000 training images \\
Calibration pairs   & 10,000 matched pairs \\
Calibration sampling & $x_0,y$ from train, $t{\sim}\mathrm{U}\{0,T{-}1\}$ \\
Calibration $x_t$   & single-step noising; shared \\
Calibration metrics & $\rho$, $\cos(\Delta)$, $\mathrm{MSE}_\Delta$ \\
Calibration aggregation & $\rho$: ratio of expectations \\
                    & $\cos(\Delta)$, $\mathrm{MSE}_\Delta$: per-sample mean \\
Cosine floor        & $10^{-8}$ \\
\bottomrule
\end{tabular}
\end{table}

Figure~\ref{fig:arch} shows the corresponding UNet architectures.
Compression applies simultaneous channel halving ($128{\to}64$) and
depth reduction ($\mathrm{nrb}\;2{\to}1$), yielding $5.9{\times}$
parameter reduction. Self-attention is retained at $16{\times}16$
resolution in both models, following the teacher's attention
configuration.

\begin{figure}[t]
\centering
\figorbox{0.93\textwidth}{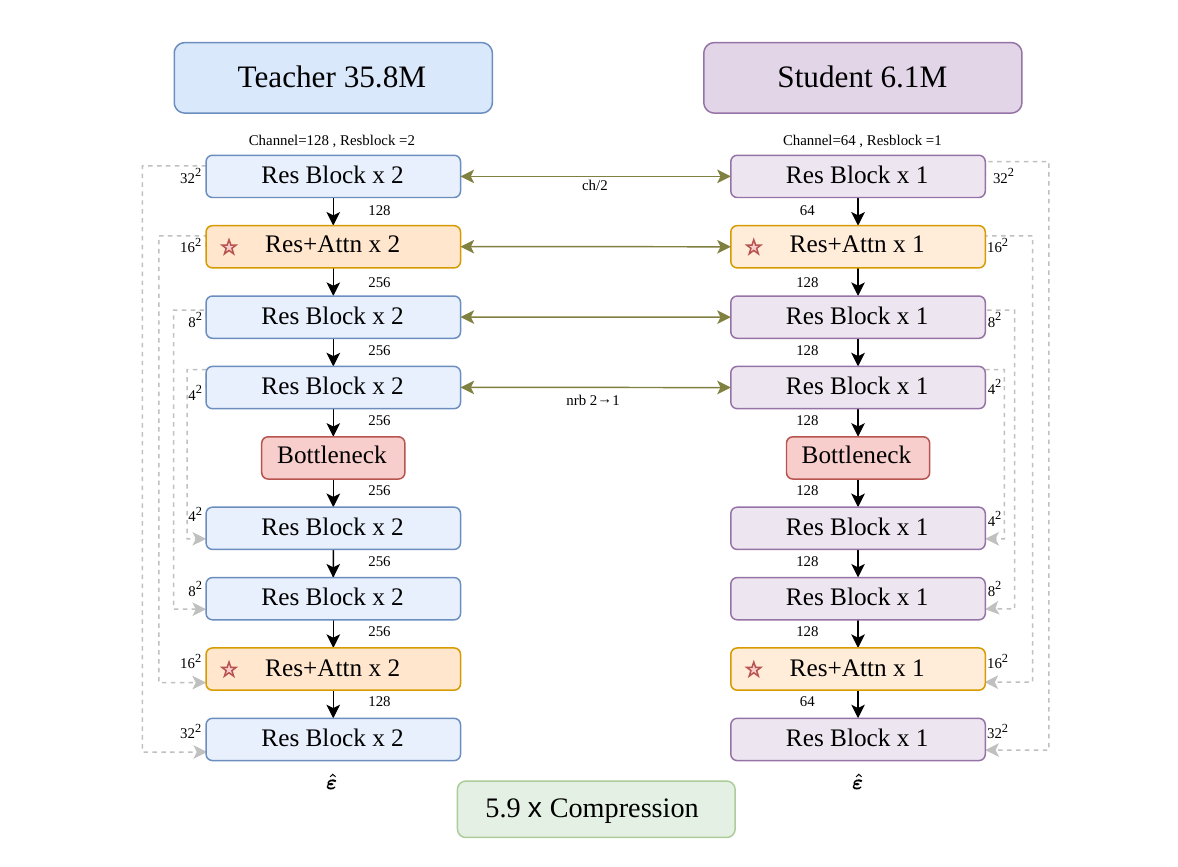}
\caption{Teacher (35.8M) and student (6.1M) UNet architectures.
Compression halves channels ($128{\to}64$) and reduces depth
($\mathrm{nrb}\;2{\to}1$). Self-attention ($\bigstar$) is retained at
$16{\times}16$ in both. Channel widths are annotated on arrows; skip
connections (dashed) operate at all four resolutions.}
\label{fig:arch}
\end{figure}

Table~\ref{tab:efficiency} reports parameters, inference passes,
FLOPs, latency, and peak memory for the teacher, \dash, and the two
compression baselines. The measured 50-step speedup ($3.62{\times}$ at batch 64)
is below the $5.77{\times}$ theoretical FLOPs ratio, consistent with fixed
kernel and per-layer overheads that a larger model amortises better.
\dash and
composite distillation share an architecture and therefore an
inference cost: the calibration advantage is orthogonal to speed,
and the efficiency gain is relative to the full teacher.

\begin{table}[t]
\caption{Inference efficiency, measured on a single T4 GPU with
identical sampler and batch settings, in fp32. \dash and pruning emit
two branches and so run two passes per step; the single-head baseline
runs one. Figures are reported at batch~64, where memory is dominated by
activations rather than by a fixed allocator floor.
$^\ast$Analytic, counted from the layer configuration; 50-step totals
use unrounded values; both compression
baselines carry more parameters than the 6.1M \dash student and use
the student width at the teacher depth
($\mathrm{ch}{=}64$, $\mathrm{nrb}{=}2$). Latency is the minimum of three timed repetitions after two warm-up
iterations, with device synchronisation before and after each
repetition; memory is peak allocation. The two compression columns
were profiled in a separate pass on the same device; re-profiling the
teacher and \dash under that pass reproduced their latency and
throughput to within roughly $15\%$, so the small \dash{}/single-head
difference is comparable to profiling variance and the
teacher-to-\dash{} speedup is the robust comparison.}
\label{tab:efficiency}
\centering\small\setlength{\tabcolsep}{8pt}
\begin{tabular}{lcccc}
\toprule
Metric & Teacher & \dash & Single-head & Magnitude pruning \\
\midrule
Parameters (M)              & 35.75   & 6.08   & 8.99  & 8.95 \\
Inference passes/step       & 2       & 2      & 1     & 2 \\
FLOPs, single pass (G)$^\ast$ & 12.48 & 2.16 & 3.19 & 3.19 \\
FLOPs, 50-step CFG (G)$^\ast$ & 1247.8 & 216.3 & 159.7 & 319.4 \\
Peak memory, bs=64 (MB)     & 821.1   & 349.9  & 419.5 & 383.6 \\
Latency, bs=64 (ms)         & 14217.3 & 3925.8 & 3424.8 & 7241.0 \\
Samples/sec, bs=64          & 4.50    & 16.30  & 18.69 & 8.84 \\
\bottomrule
\end{tabular}
\end{table}

\subsection{Reproducibility Details}
\label{app:repro}

Table~\ref{tab:repro_env} records the environment, data pipeline and
evaluation implementation, Table~\ref{tab:repro_arch} the conditioning
path shared by both models, and Table~\ref{tab:repro_baselines} one
configuration block per baseline.

Two entries carry more weight than the rest. FID values are tied to one
library, one feature network and one reference subset, so they are
internally comparable but not to figures computed under a different
implementation. And the calibration metrics use a \emph{shared} $x_t$
per pair, so $\rho$ and $\cos(\Delta)$ measure a difference between
models rather than between noise draws.

\begin{table}[t]
\caption{Environment, data pipeline, and evaluation implementation.}
\label{tab:repro_env}
\centering\small\setlength{\tabcolsep}{3pt}
\begin{tabular}{@{}ll@{}}
\toprule
Setting & Value \\
\midrule
\multicolumn{2}{@{}l}{\textit{CIFAR-10\,/\,-100 training}} \\
Accelerator          & Single NVIDIA T4 16\,GB \\
Framework            & PyTorch 2.0 \\
Numerical precision  & mixed (\texttt{torch.autocast}) \\
Wall-clock per run   & 24\,/\,25\,h teacher; 16\,/\,17\,h student \\
\midrule
\multicolumn{2}{@{}l}{\textit{ImageNet-64 training}} \\
Accelerator          & TPU v5e-8 \\
Chips / cores        & 8 chips, 1 core each \\
Topology             & $2{\times}4$ slice, 2D torus \\
Framework            & PyTorch 2.0, \texttt{torch\_xla} \\
Numerical precision  & fp32 params, default \textsc{mxu} matmul \\
Wall-clock per run   & 115\,h teacher; 72\,h student \\
\midrule
\multicolumn{2}{@{}l}{\textit{Data pipeline}} \\
CIFAR source         & \texttt{torchvision} train (50{,}000) \\
ImageNet-64 source   & Downsampled ImageNet (1{,}281{,}167) \\
Preprocessing        & \texttt{uint8}$\to$float, to $[-1,1]$ \\
Augmentation         & none (same pipeline for all) \\
Class sampling       & uniform at generation \\
Shuffling seed       & per-run seed (42\,/\,123\,/\,456) \\
\midrule
\multicolumn{2}{@{}l}{\textit{FID\,/\,IS implementation}} \\
Library and version  & \texttt{torch-fidelity} 0.4.0 \\
Feature network      & InceptionV3 (library default) \\
Input preprocessing  & 8-bit PNG; default resize \\
Reference statistics & 50{,}000 training images \\
Reference subset seed & 0 \\
Covariance precision & library default (float64) \\
IS split count       & library default (10) \\
\midrule
\multicolumn{2}{@{}l}{\textit{Checkpointing}} \\
Selection rule       & final epoch; no validation \\
Teacher weights      & EMA \\
\bottomrule
\end{tabular}
\end{table}

\begin{table}[t]
\caption{Conditioning and architecture details not in
Table~\ref{tab:arch_train}. Both models use the same conditioning path.}
\label{tab:repro_arch}
\centering\small\setlength{\tabcolsep}{2pt}
\begin{tabular}{@{}ll@{}}
\toprule
Setting & Value \\
\midrule
Timestep embedding & sinusoidal, width $4{\times}$ base ch. \\
Class conditioning & learned embedding, added to $t$-embedding \\
Null token         & one extra embedding row, index $C$ \\
Dropout (arch.)    & $0.0$ \\
Normalisation      & GroupNorm, 32 groups \\
Skip connections   & concatenation, one per encoder block \\
Resampling         & strided conv down, nearest$+$conv up \\
\bottomrule
\end{tabular}
\end{table}

\begin{table}[t]
\caption{Per-baseline configuration. The 6.1M students share the
234K-iteration budget, the frozen teacher, and frozen TIRT Transfer
weights; \textit{Scratch} uses no teacher and therefore no TIRT
Transfer, and the two compression baselines differ in size. Inference passes are the
forward passes required per denoising step at evaluation. ($^\S$) Branch selected by the teacher's
own $10\%$ null-token dropout; \textit{Scratch} trains with the same
$10\%$ rate so that a null branch exists at all. Magnitude pruning
ranks channels by $\ell_1$ norm over eight iterative steps, leaving the
output convolution and embeddings unpruned; counts are rounded to
multiples of $32$ (GroupNorm-$32$), which is why the realised size is
8.95M rather than the 6.1M target. FitNets matches features at all four encoder and decoder resolutions,
where teacher and student spatial sizes coincide, and optimises the
hint loss jointly with the composite objective rather than as the
separate pre-training stage of the original formulation, so that it differs from \textit{Composite} in exactly one
term. Single-head embeds $w$ as a $128$-dimensional Fourier feature
added to the timestep embedding and has no null-token row; it reproduces the guidance-distillation stage
of Meng~\etal~\cite{meng2023distill} only, not their subsequent
progressive step distillation.}
\label{tab:repro_baselines}
\centering\small\setlength{\tabcolsep}{4pt}
\resizebox{\textwidth}{!}{%
\begin{tabular}{lccccc}
\toprule
Baseline & Params & Teacher queries & Training $w$ & Loss & Inference passes \\
\midrule
Scratch                  & 6.1M  & 0 & --      & ground-truth $\eps$          & 2 \\
Cond-only                & 6.1M  & 1 & --      & $\|\eps^y_S{-}\eps^y_T\|^2$  & 2 \\
Composite                & 6.1M  & 2 & $4.0$   & $\|\tilde\eps_S{-}\tilde\eps_T\|^2$ & 2 \\
Multi-$w$ composite      & 6.1M  & 2 & $\mathrm{U}[1.5,5.0]$ & $\|\tilde\eps_S{-}\tilde\eps_T\|^2$ & 2 \\
FitNets                  & 6.1M  & 2 & $4.0$   & composite $+\,\lambda_{\mathrm{feat}}$ feature MSE, $\lambda_{\mathrm{feat}}{=}1$ & 2 \\
Gap $\Delta$ match       & 6.1M  & 2 & --      & $\|\Delta_S{-}\Delta_T\|^2$  & 2 \\
Label-dropout$^\S$       & 6.1M  & 1 & --      & $\|\eps^{y\,\text{or}\,\varnothing}_S{-}\eps^{y\,\text{or}\,\varnothing}_T\|^2$ & 2 \\
Magnitude pruning        & 8.95M & 0 & --      & ground-truth $\eps$ (finetune) & 2 \\
Single-head, $w$-cond.   & 8.99M & 2 & $\mathrm{U}[1.5,5.0]$ & $\|\tilde\eps_S(w){-}\tilde\eps_T(w)\|^2$ & 1 \\
\midrule
\dash~(ours)             & 6.1M  & 2 & --      & $\Ltot$                      & 2 \\
\bottomrule
\end{tabular}}
\end{table}

\section{Extended Related Work}
\label{app:related}

Table~\ref{tab:rw} positions \dash against the closest prior methods
across four axes: whether the method targets parameter compression,
step reduction, explicit branch-level calibration, and curriculum
transfer from teacher to student. Parameter compression and step
reduction are orthogonal efficiency axes;
\dash is the only method in this comparison that addresses
compression while also constraining the guidance branches directly.

Among these, Meng~\etal~\cite{meng2023distill} are closest in
objective: they distil classifier-free guidance into a single
$w$-conditioned student, removing CFG's second forward pass at
teacher-scale capacity. The branch decomposition need not be
recovered in that regime, and they report the single-head
formulation preferable there. Under $5.9{\times}$ compression the
composite baseline instantiates the same objective on a compressed
student, where the non-identifiability characterised in
\S\ref{app:proof_underdet} becomes
consequential ($\rho{=}0.68$ versus $0.91$ for \dash). The two
axes remain composable: a \dash-compressed student can subsequently
undergo step reduction~\cite{progdistill}.

\noindent\textbf{Method details.}
The main text compresses the following descriptions for space.

\noindent\emph{Step reduction.} Luhman~\etal~\cite{luhman2021} introduce
student--teacher distillation for diffusion; progressive
distillation~\cite{progdistill} and consistency models~\cite{cm,lcm}
shorten the sampling trajectory, while solver-based
methods~\cite{dpmsolver,dpm_plus} and adversarial or
distribution-matching objectives~\cite{add,dmd} reach few-step sampling
by other means. Distribution matching distillation~\cite{dmd} aligns
the student's output distribution with the teacher's, again without
targeting parameter count. Adapter-based guidance
distillation~\cite{agd2025} approximates the composite guided
prediction with lightweight adapters on a frozen base model, halving
inference cost while leaving the base parameter count unchanged.

\noindent\emph{Parameter compression.} BK-SDM~\cite{bksdm} removes UNet blocks
from Stable Diffusion and SnapFusion~\cite{snapfusion} targets mobile
deployment; both operate at 512px in latent space with text
conditioning, so neither exposes the class-conditional pixel-space
branch structure studied here. Diff-Pruning~\cite{diffpruning} prunes a
trained teacher by Taylor-based importance and finetunes. Taylor
importance is defined against a single training loss; on a
classifier-free-guided model the gradient may be taken through the
conditional branch, the unconditional branch, or their composite, and
these rank filters differently. That choice is the branch ambiguity
this paper isolates, so the pruning baseline adopts the magnitude criterion, which requires
no such choice and is therefore generic rather than a Diff-Pruning
reproduction.

\noindent\emph{Theoretical analyses of composite objectives.} Score
distillation via reparametrised DDIM~\cite{distilldirect} analyses
composite guided objectives theoretically but does not address the
branch-level non-identifiability identified here. Concurrent
work~\cite{li2026rethinking} identifies a related under-identification
in on-policy CFG distillation; that failure is dynamic, arising from
privileged negative-branch information available to the teacher but
not the student during student-generated rollouts, and is distinct
from the static null-space characterisation proven in
\S\ref{app:proof_underdet} for offline compression.

\noindent\emph{Feature-level distillation.} FitNets~\cite{fitnets} aligns hidden
feature maps between teacher and student, underpinning much of the
subsequent knowledge-distillation literature. The main text adopts it
as a baseline testing whether richer internal supervision, added to
composite supervision, can substitute for a constraint on the emitted
score branches.

\noindent\emph{Sample-axis reweighting.} DoReMi~\cite{doremi},
InfoBatch~\cite{infobatch}, and RegMix~\cite{regmix} reweight or
resample training examples for large language models, not diffusion
models specifically. Within diffusion training,
Li~\etal~\cite{li2025pruning} prune and reweight the training set
jointly, and Kim~\etal~\cite{kim2024unbiased} correct for dataset
bias via importance reweighting of samples. All five select
\emph{which examples} to prioritise; TIRT instead reweights
\emph{timesteps} within every example identically, the orthogonal
axis noted in the main text.

\begin{table}[t]
\caption{Comparison of distillation approaches. \cmark~=~addressed;
\xmark~=~not addressed. Parameter compression denotes reducing the
denoising network parameter count relative to the teacher; step
reduction denotes fewer denoising steps, and CFG-pass reduction fewer
guidance forward passes per step. These axes are orthogonal. \dash targets compression and is
composable with step-reduction methods~\cite{progdistill,cm}.}
\label{tab:rw}
\centering\small\setlength{\tabcolsep}{14pt}
\resizebox{\textwidth}{!}{%
\begin{tabular}{lccccc}
\toprule
Method &
  \makecell{Param\\Compr.} &
  \makecell{Step\\Reduce} &
  \makecell{CFG-pass\\Reduce} &
  \makecell{Branch\\Calib.} &
  \makecell{Curric.\\Transfer} \\
\midrule
Luhman~\etal~\cite{luhman2021}    & \xmark & \cmark & \xmark & \xmark & \xmark \\
Meng~\etal~\cite{meng2023distill} & \xmark & \cmark & \cmark & \xmark & \xmark \\
AGD~\cite{agd2025}                & \xmark & \xmark & \cmark & \xmark & \xmark \\
Prog.\ Distil.~\cite{progdistill} & \xmark & \cmark & \xmark & \xmark & \xmark \\
BK-SDM~\cite{bksdm}               & \cmark & \xmark & \xmark & \xmark & \xmark \\
SnapFusion~\cite{snapfusion}      & \cmark & \cmark & \xmark & \xmark & \xmark \\
BOOT~\cite{boot}                  & \xmark & \cmark & \xmark & \xmark & \xmark \\
Diff-Pruning~\cite{diffpruning}   & \cmark & \xmark & \xmark & \xmark & \xmark \\
\dash~(ours)                      & \cmark & \xmark & \xmark & \cmark & \cmark \\
\bottomrule
\end{tabular}}
\end{table}


\section{Theoretical Analysis}
\label{app:theory}

The following results formalise the core claims of \dash for fixed
input $(\xt,t,y)$ with deterministic outputs; the stochastic case
follows by linearity of expectation.

\subsection{Solution Geometry of Composite Distillation}
\label{app:proof_underdet}

Define per-branch residuals
$\delta^y \triangleq \eps^y_S - \eps^y_T$ and
$\delta^\varnothing \triangleq \eps^\varnothing_S -
\eps^\varnothing_T$.
Using the CFG definition, the composite residual is
\begin{equation}
\tilde\eps_S - \tilde\eps_T
= (1-w)\,\delta^\varnothing + w\,\delta^y.
\label{eq:comp_residual}
\end{equation}
Setting $\mathcal{L}_{\mathrm{comp}}{=}0$ requires this to vanish,
which constrains only the weighted sum of residuals.
The zero-loss surface is therefore the null space
\begin{equation}
\mathcal{N} =
\bigl\{(\delta^y,\,\delta^\varnothing)\in\mathbb{R}^d
\times\mathbb{R}^d : w\,\delta^y + (1{-}w)\,\delta^\varnothing
= \mathbf{0}\bigr\},
\label{eq:null}
\end{equation}
a $d$-dimensional linear subspace for every $w$: the coefficient
operator $[\,w\mathbf{I}\;|\;(1{-}w)\mathbf{I}\,]$ has full rank
$d$ unless $w{=}0$ and $1{-}w{=}0$ simultaneously, which cannot
occur.
Guidance collapse ($\Delta_S{=}0$, \ie
$\eps^y_S{=}\eps^\varnothing_S$) is achieved by setting both
branches equal to the composite teacher prediction:
$\eps^y_S{=}\eps^\varnothing_S{=}\tilde\eps_T$.
This gives residuals
$\delta^y{=}\tilde\eps_T - \eps^y_T$ and
$\delta^\varnothing{=}\tilde\eps_T - \eps^\varnothing_T$,
which satisfy~\eqref{eq:null}:
\begin{align}
w\,\delta^y &+ (1{-}w)\,\delta^\varnothing \nonumber\\
&= w(\tilde\eps_T - \eps^y_T)
   + (1{-}w)(\tilde\eps_T - \eps^\varnothing_T) \nonumber\\
&= \tilde\eps_T
   - \bigl[w\,\eps^y_T + (1{-}w)\,\eps^\varnothing_T\bigr]
 = \tilde\eps_T - \tilde\eps_T = \mathbf{0}.
\label{eq:collapse_verify}
\end{align}
This establishes guidance collapse as a \emph{global minimiser} of
$\mathcal{L}_{\mathrm{comp}}$, not a local saddle point or
initialisation artefact.

\vspace{4pt}
\noindent\textbf{Proposition~C.1} (Exact invariance in the collapse
direction)\textbf{.} For a fixed input and any $u\in\mathbb{R}^d$,
transform the branch residuals as $\delta^y\to\delta^y+(1{-}w)u$ and
$\delta^\varnothing\to\delta^\varnothing-wu$. The composite residual
is unchanged,
\begin{equation}
w\bigl(\delta^y{+}(1{-}w)u\bigr)+(1{-}w)\bigl(\delta^\varnothing{-}wu\bigr)
= w\delta^y+(1{-}w)\delta^\varnothing,
\label{eq:invariance}
\end{equation}
so $\mathcal{L}_{\mathrm{comp}}$ is exactly invariant along this
branch-output direction at \emph{every} residual value, not merely
where the loss vanishes. The non-identifiability therefore does not
depend on the student reaching zero loss. This is a statement about
the two emitted branch outputs at a fixed input; because the branches
share parameters across all inputs, it does not by itself establish
that the parameter loss has the same flat direction.

\vspace{4pt}
\noindent\textbf{Where composite solutions land.} The residuals
$\delta^y{=}\eps^y_S{-}\eps^y_T$ and
$\delta^\varnothing{=}\eps^\varnothing_S{-}\eps^\varnothing_T$ give
two checks on~\eqref{eq:null} that require no free parameters. First,
every point of the null space satisfies
$\|\delta^y\|/\|\delta^\varnothing\|{=}(w{-}1)/w$, which is $0.75$
at $w{=}4$ regardless of position along the line. Second, a solution lying a
fraction $f{=}\langle\delta^\varnothing,\Delta_T\rangle/(w\|\Delta_T\|^2)$
of the way toward collapse ($\delta^\varnothing{=}fw\Delta_T$) has
$\Delta_S{=}\Delta_T{+}\delta^y{-}\delta^\varnothing{=}(1{-}f)\Delta_T$,
so $\rho{=}1{-}f$ exactly on the line and $f{\geq}1{-}\rho$ in
general, since any component of $\delta^\varnothing$ orthogonal to the
line inflates $\|\Delta_S\|$. Measured on the composite students:

\vspace{2pt}
\centerline{\small
\begin{tabular}{lccc}
\toprule
& $\|\delta^y\|/\|\delta^\varnothing\|$ & $f$ & $1{-}\rho$ \\
\midrule
CIFAR-10    & 0.77 & 0.34 & 0.32 \\
CIFAR-100   & 0.73 & 0.38 & 0.35 \\
ImageNet-64 & 0.79 & 0.40 & 0.35 \\
\bottomrule
\end{tabular}}
\vspace{2pt}

\noindent The ratio scatters on both sides of $0.75$ (mean $0.763$,
worst $5.3\%$), and $f{\geq}1{-}\rho$ holds on all three with the
excess growing as the task gets harder, consistent with the orthogonal
term above.

\vspace{4pt}
\noindent\textbf{Remark~C.2} (Determinacy under dual
supervision)\textbf{.}
The losses $\Lim{=}\mathbb{E}[\hat\omega_t\|\delta^y\|^2]$ and
$\Lun{=}\mathbb{E}[\hat\omega_t\|\delta^\varnothing\|^2]$ constrain
$\delta^y{=}\mathbf{0}$ and $\delta^\varnothing{=}\mathbf{0}$
independently, reducing $\mathcal{N}$ to a single point and
excluding all degenerate solutions in~\eqref{eq:null}, since
$\hat\omega_t{>}0$ strictly ($\hat\omega_t$ is the unit-mean
normalisation of $\omega_t{=}[\min(\snr_t,\gamma)/\snr_t]\cdot\sigma(\lambda_t)$
defined in the main text, and $\sigma(\cdot)$ is strictly positive),
so neither loss vanishes without its residual being exactly zero.

\vspace{4pt}
\noindent\textbf{Corollary~C.3} (Determinacy under two-weight
supervision)\textbf{.}
The null space~\eqref{eq:null} is defined at a single guidance
weight $w$. If composite supervision were instead applied at two
distinct weights $w_1{\neq}w_2$, zero loss at both requires
$w_1\delta^y{+}(1{-}w_1)\delta^\varnothing{=}\mathbf{0}$ and
$w_2\delta^y{+}(1{-}w_2)\delta^\varnothing{=}\mathbf{0}$
simultaneously; subtracting gives
$(w_1{-}w_2)(\delta^y{-}\delta^\varnothing){=}\mathbf{0}$, so
$\delta^y{=}\delta^\varnothing$, and substituting back forces
$\delta^y{=}\delta^\varnothing{=}\mathbf{0}$. Two distinct
guidance weights therefore determine both branches exactly, in the
noiseless limit. This does not make dual-branch supervision
redundant. Multi-$w$ composite training obtains this determinacy only
in expectation across batches, through a linear mixture of targets at
each individual step. Supervision at two fixed weights is exact on
every sample, but is a linearly reparametrised form of
$(\Lim,\Lun)$ whose conditioning depends on the separation
$|w_2{-}w_1|$ (Remark~C.4).

\smallskip
\noindent\textbf{Remark~C.4} (Conditioning)\textbf{.}
Writing $\tilde\eps(w){=}(1{-}w)\eps^\varnothing{+}w\eps^y$,
supervision at two weights is the linear map
$M{=}\bigl[\begin{smallmatrix}1{-}w_1 & w_1\\ 1{-}w_2 &
w_2\end{smallmatrix}\bigr]$ with $\det M{=}w_2{-}w_1$, so its
zero-loss set is identical to that of $(\Lim,\Lun)$. The Gram matrix
$M^\top M$ is not: its condition number is $152$ at
$(w_1,w_2){=}(1.5,5.0)$ and diverges as $w_2{\to}w_1$
($6.3{\times}10^4$ at $(3.9,4.1)$). These are deterministic two-weight
figures and are not comparable to the ${\approx}304$ quoted in the main
paper, which is the condition number of the \emph{expected} Gram matrix
$\mathbb{E}[a a^\top]$ under $w{\sim}\mathrm{U}[1.5,5.0]$, the
multi-$w$ baseline's sampling distribution. At $(w_1,w_2){=}(0,1)$,
$M{=}I$:
$\Lim$ and $\Lun$ are the unit-conditioned member of this family: the
basis in which the two residuals are the branch errors themselves.

\vspace{4pt}
\noindent\textbf{Remark~C.5} (Inference-time scalar rescaling)\textbf{.} Rescaling
the student gap by a constant $a$, equivalently sampling at $w'{=}aw$,
minimises $\mathbb{E}[\tfrac{1}{d}\|a\Delta_S{-}\Delta_T\|^2]$ at
\begin{equation}
a^{\ast}=\frac{\mathbb{E}[\Delta_S^{\!\top}\Delta_T]}{\mathbb{E}[\|\Delta_S\|^2]},
\label{eq:rescale}
\end{equation}
which depends on second moments of the matched pairs. Because $\rho$ is a
ratio of expected norms and $\cos(\Delta)$ a mean of per-sample cosines,
$a^{\ast}$ is not recoverable from those two summaries alone. No choice of $a$ rotates
$\Delta_S$, so the directional component of the residual survives.

\subsection{Curriculum Gradient Misalignment Under Distillation}
\label{app:proof_tirt}

Let $e_t$ denote the relevant per-timestep training error and let
$\omega_t{=}a_t\sigma(\lambda_t)$, where
$a_t{=}\min(\snr_t,\gamma)/\snr_t$. With the unit-mean normalisation
defined in the main text, the weighted objective is proportional to
\begin{equation}
\mathcal{L} = \frac{\sum_j \omega_j e_j}{\sum_j \omega_j}.
\label{eq:tirt_obj}
\end{equation}
Holding the network outputs fixed, the quotient rule gives
\begin{equation}
\frac{\partial\mathcal{L}}{\partial\lambda_t}
= \frac{a_t\,\sigma'(\lambda_t)}{\sum_j \omega_j}\bigl(e_t - \mathcal{L}\bigr).
\label{eq:tirt_grad}
\end{equation}
The schedule update at a timestep therefore depends on that timestep's
error relative to the current weighted average error, not on its raw
value.

Two consequences follow. First, teacher training uses ground-truth
noise-prediction errors, whereas \dash{} distillation is dominated by
student--teacher branch-matching errors plus the anchor term. These
error profiles are generally different, so continuing to optimise
$\lambda_t$ during distillation need not preserve the teacher's final
learned weighting. \dash{} therefore treats the teacher's final
normalised schedule as a fixed prior.

Second, \eqref{eq:tirt_grad} is not by itself a difficulty-selection
argument: for network outputs held fixed, gradient descent moves
$\lambda_t$ \emph{down} where $e_t{>}\mathcal{L}$. In training,
$\lambda$ and the network parameters co-adapt, so the joint fixed point
is not the static one; $\sigma(\cdot)$ bounds the modulation and the
unit-mean normalisation fixes its total. The paper therefore reports
the learned schedule and its measured effect rather than a mechanism:
the schedule's measured benefit over fixed Min-SNR weighting is given
in Table~\ref{tab:teacher}, and freezing it for the student is
preferred to relearning from $\lambda^S{=}0$ within the tested
234K-iteration budget, by $3.84$\,/\,$3.94$ FID
(Table~\ref{tab:ablation_std}).

\section{Extended Ablations}
\label{app:ablations}
\subsection{Teacher Component Ablation}
\label{app:teacher_ablation}

Table~\ref{tab:teacher} decomposes the teacher's gain over a Min-SNR
baseline into TIRT and TAG. On CIFAR-10 the two contribute
2.05 and 1.79 FID points respectively; their combined gain (3.74) is
near-additive relative to the sum (3.84), with $2.6\%$ overlap. On
CIFAR-100 the combined gain (4.04) is moderately subadditive
relative to the sum (4.60), with $12.2\%$ overlap, reflecting
partial interaction between the two components at 100 fine-grained
classes. TIRT is evaluated at fixed $w{=}1.5$ ($\dagger$) to isolate training
effects, since TAG's adaptive inference schedule cannot be held fixed
independently of its own effect. The fixed-guidance row
($^{\ddagger}$) separates the two sources of TAG's benefit: $65\%$
derives from raising $w$ from 1.5 to 4.0 (1.16 FID points), and $35\%$
from the adaptive sigmoid schedule (0.63 points).

\begin{table}[t]
\caption{Teacher component ablation. FID$\downarrow$\,/\,IS$\uparrow$,
50K samples, 50-step DDIM\@. ($\dagger$) fixed $w{=}1.5$ to isolate
training effects. ($\ddagger$) fixed $w{=}4.0$, no adaptive schedule.
Published rows are provided for scale only: they use their original
architectures, sampling budgets (Steps column), training protocols and
evaluation implementations, and are therefore not controlled
comparisons with the models trained here, which use 50 steps without
augmentation.}
\label{tab:teacher}
\centering\small\setlength{\tabcolsep}{14pt}
\resizebox{\textwidth}{!}{%
\begin{tabular}{l cc cc cc}
\toprule
& & & \multicolumn{2}{c}{CIFAR-10} & \multicolumn{2}{c}{CIFAR-100} \\
\cmidrule(lr){4-5}\cmidrule(lr){6-7}
Method & Params & Steps & FID$\downarrow$ & IS$\uparrow$
  & FID$\downarrow$ & IS$\uparrow$ \\
\midrule
DDPM~\cite{ddpm}   & 35M  & 1000 & 3.17 & 9.46 & 4.21 & 7.58 \\
iDDPM~\cite{iddpm} & 53M & 4000 & 2.90 & 9.58 & 3.64 & 7.82 \\
EDM~\cite{edm}     & 56M &   35 & 1.97 & 9.68 & 2.44 & 8.12 \\
\midrule
\multicolumn{7}{l}{\textit{\dash{} teacher (35.8M, 50 steps):}} \\
Min-SNR base$^{\dagger}$  & 35.8M & 50 & 9.21 & 8.54 & 10.84 & 7.18 \\
\;+Fixed CFG$^{\ddagger}$ & 35.8M & 50 & 8.05 & 8.79 &  9.46 & 7.38 \\
\;+TAG only               & 35.8M & 50 & 7.42 & 8.94 &  8.71 & 7.43 \\
\;+TIRT only$^{\dagger}$  & 35.8M & 50 & 7.16 & 8.71 &  8.37 & 7.29 \\
\;Full (TAG+TIRT)         & 35.8M & 50 & \textbf{5.47} & \textbf{9.42}
                                       & \textbf{6.80} & \textbf{7.46} \\
\bottomrule
\end{tabular}}
\end{table}

\subsection{Teacher Design Schedules}
\label{app:teacher_design}

Figure~\ref{fig:teacher_design} shows the two schedules governing
teacher training and inference. TAG modulates the guidance weight
across the denoising trajectory by the sigmoid schedule of
\S\ref{sec:method}, with $w_{\min}{=}1.0$, $\kappa{=}5.0$, $w_{\max}{=}4.0$
(Table~\ref{tab:arch_train}), giving $w{\approx}1$ at high noise and
$w{\approx}4$ at low noise. TIRT concentrates timestep weight at
$t{=}400$--$700$ with a $2.3{\times}$ peak-to-valley ratio in the learned modulation
$\sigma(\lambda_t)$ relative to the Min-SNR baseline.

\begin{figure}[t]
\centering
\setlength{\tabcolsep}{3pt}
\resizebox{\textwidth}{!}{%
\begin{tabular}{@{}cc@{}}
\figorbox{0.46\textwidth}{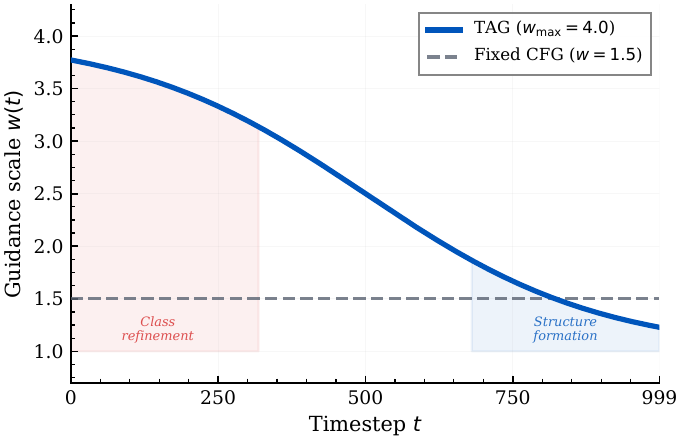} &
\figorbox{0.46\textwidth}{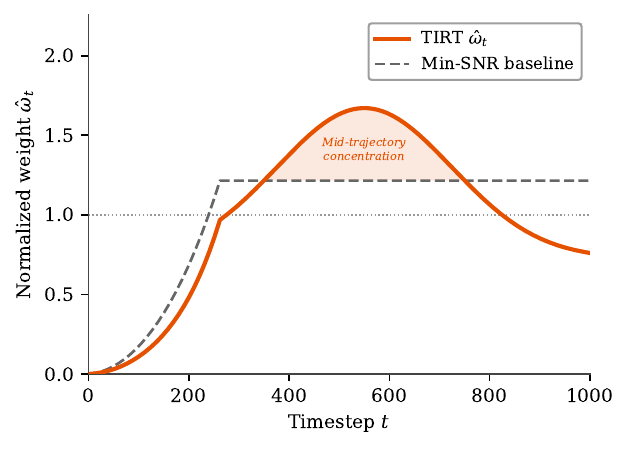} \\[2mm]
{\small (a)~TAG: $w(t)$} & {\small (b)~TIRT: $\hat\omega_t$}
\end{tabular}}
\caption{Teacher design schedules (CIFAR-10). (a)~TAG applies
asymmetric guidance across the noise trajectory. (b)~TIRT
concentrates training weight at mid-trajectory timesteps, where
the base schedule is flat.}
\label{fig:teacher_design}
\end{figure}

\noindent\textbf{TIRT Transfer and the compression baselines.}
Magnitude pruning and single-head receive no TIRT Transfer. Crediting both the
$2.60$\,/\,$2.67$ FID it contributes (Table~\ref{tab:ablation}) still leaves
single-head $5.31$\,/\,$10.36$ behind \dash and pruning
$6.57$\,/\,$11.34$.

\subsection{Standard Deviations Across Seeds}
\label{app:std}
Baseline and ablation configuration names (Cond-only, Composite, Label-dropout,
Gap match, Multi-$w$ composite, FitNets, Magnitude pruning, Single-head,
w/o~$\mathcal{L}_{\cdot}$)
are defined in the main text's Setup and Ablation sections; this and
the following subsections reuse those names without redefining them.
Tables~\ref{tab:distill_std} and~\ref{tab:ablation_std} reproduce
the main results with mean~$\pm$~std, computed over three independent
training runs (seeds 42\,/\,123\,/\,456). The seed controls
initialisation, data order and the CFG dropout mask, so the spread
covers training as well as sampling. Variance is uniformly low for the
full \dash model. Configurations that remove unconditional supervision
(Cond-only, w/o~\Lun) show elevated variance relative to the full
model ($\pm0.46$ and $\pm0.31$ against $\pm0.12$), consistent with the
null-space geometry in \S\ref{app:proof_underdet}: with no explicit
target for $\eps^\varnothing_S$, which point of the null space a run
reaches depends on initialisation and data order.

\noindent\textbf{The two unsupervised-branch configurations.}
Cond-only and w/o~\Lun both leave $\eps^\varnothing_S$ without a
target, and both collapse ($\rho{=}0.09$ and $0.23$). The $3.47$ FID
between them comes from \Lan alone, against $1.67$ when \Lan is
removed from the full model: with the gap inert the sampler
effectively follows the conditional branch, so grounding that branch
in ground-truth noise transfers directly to sample quality while IS is
unchanged ($7.21{\to}7.23$). \textit{Scratch} completes the picture from a third
direction. Training with $10\%$ null-token dropout and no teacher
(Table~\ref{tab:repro_baselines}), it holds a live gap
($\rho{=}0.80$\,/\,$0.78$) pointing largely independently of the
teacher's ($\cos(\Delta){=}0.51$\,/\,$0.48$): a $0.29$\,/\,$0.30$
excess of $\rho$ over $\cos(\Delta)$, where no distilled configuration
with a defined $\cos(\Delta)$ exceeds $0.05$. Its
$\mathrm{MSE}_\Delta$ ($0.213$) nearly coincides with w/o~\Lun's
($0.221$) although $\rho$ differs by $0.57$, so a largely intact but
misdirected gap and a near-inert one are indistinguishable on that
statistic alone.
Comparing w/o~\Lun with w/o~\Lim runs the other way: w/o~\Lun is
worse on $\rho$ ($0.23$ versus $0.41$) and $\cos(\Delta)$ ($0.27$
versus $0.46$) yet better on FID and $\mathrm{MSE}_\Delta$. A
near-inert gap costs less at sampling than a live but misdirected one,
which $w$ amplifies at every denoising step;
$\mathrm{MSE}_\Delta$ separates the two where $\rho$ alone does not.
The removal cost of \Lun is non-monotone in class count
($9.97$\,/\,$12.48$\,/\,$11.03$ FID), so the three are reported
separately rather than as a trend in label-space size.

\begin{table}[t]
\caption{CIFAR distillation results, mean$\pm$std over three
independent training runs (seeds 42\,/\,123\,/\,456); the teacher row
is the spread of three teacher runs, and all students are distilled
from a single frozen teacher checkpoint. Loss-formulation rows share the ch=64, 6.1M
student; magnitude pruning is 8.95M and single-head 8.99M. FID$\downarrow$\,/\,IS$\uparrow$, 50K samples,
50-step DDIM, TAG $w_{\max}{=}4.0$. ImageNet-64 in Table~\ref{tab:in64_std}.}
\label{tab:distill_std}
\centering\small\setlength{\tabcolsep}{4pt}
\begin{tabular}{lcccc}
\toprule
Model &
  \makecell{C10\\FID$\downarrow$} &
  \makecell{C10\\IS$\uparrow$} &
  \makecell{C100\\FID$\downarrow$} &
  \makecell{C100\\IS$\uparrow$} \\
\midrule
Teacher
  & $5.47{\pm}0.04$ & $9.42{\pm}0.05$
  & $6.80{\pm}0.05$ & $7.46{\pm}0.04$ \\
\midrule
Scratch
  & $20.47{\pm}0.31$ & $7.44{\pm}0.09$
  & $26.14{\pm}0.44$ & $6.51{\pm}0.08$ \\
Cond-only
  & $22.31{\pm}0.46$ & $7.21{\pm}0.11$
  & $27.83{\pm}0.59$ & $5.98{\pm}0.10$ \\
Composite
  & $13.84{\pm}0.22$ & $8.63{\pm}0.08$
  & $20.84{\pm}0.37$ & $6.84{\pm}0.07$ \\
Multi-$w$ composite
  & $13.67{\pm}0.19$ & $8.96{\pm}0.09$
  & $19.13{\pm}0.37$ & $7.23{\pm}0.05$ \\
FitNets
  & $12.84{\pm}0.21$ & $8.74{\pm}0.08$
  & $19.47{\pm}0.35$ & $6.89{\pm}0.07$ \\
Gap $\Delta$ match
  & $11.42{\pm}0.18$ & $8.34{\pm}0.07$
  & $18.46{\pm}0.33$ & $6.58{\pm}0.07$ \\
Label-dropout
  & $11.43{\pm}0.11$ & $8.88{\pm}0.09$
  & $14.56{\pm}0.27$ & $7.10{\pm}0.11$ \\
Magnitude pruning
  & $18.04{\pm}0.33$ & $8.56{\pm}0.07$
  & $24.48{\pm}0.43$ & $6.89{\pm}0.08$ \\
Single-head, $w$-cond.
  & $16.78{\pm}0.25$ & $9.00{\pm}0.06$
  & $23.50{\pm}0.47$ & $7.19{\pm}0.09$ \\
\dash~(full)
  & $\mathbf{8.87{\pm}0.12}$ & $\mathbf{9.31{\pm}0.06}$
  & $\mathbf{10.47{\pm}0.16}$ & $\mathbf{7.41{\pm}0.05}$ \\
\bottomrule
\end{tabular}
\end{table}

\begin{table}[t]
\caption{ImageNet-64, mean$\pm$std over three independent training
runs (seeds 42\,/\,123\,/\,456). Same protocol as
Table~\ref{tab:distill_std}.}
\label{tab:in64_std}
\centering\small\setlength{\tabcolsep}{4pt}
\begin{tabular}{lcc}
\toprule
Model & FID$\downarrow$ & IS$\uparrow$ \\
\midrule
Teacher & $6.34{\pm}0.09$ & $48.35{\pm}0.31$ \\
\midrule
Composite & $18.43{\pm}0.43$ & $43.67{\pm}0.35$ \\
w/o \Lun & $20.13{\pm}0.31$ & $41.67{\pm}0.53$ \\
\dash~(ours) & $\mathbf{9.10{\pm}0.14}$ & $\mathbf{47.32{\pm}0.37}$ \\
\bottomrule
\end{tabular}
\end{table}

\begin{table}[t]
\caption{Component ablation, mean$\pm$std over three independent
training runs (seeds 42\,/\,123\,/\,456). Same
protocol as Table~\ref{tab:distill_std}.}
\label{tab:ablation_std}
\centering\small\setlength{\tabcolsep}{4pt}
\begin{tabular}{lcccc}
\toprule
Configuration &
  \makecell{C10\\FID$\downarrow$} &
  \makecell{C10\\IS$\uparrow$} &
  \makecell{C100\\FID$\downarrow$} &
  \makecell{C100\\IS$\uparrow$} \\
\midrule
Scratch
  & $20.47{\pm}0.31$ & $7.44{\pm}0.09$
  & $26.14{\pm}0.44$ & $6.51{\pm}0.08$ \\
w/o \Lun
  & $18.84{\pm}0.31$ & $7.23{\pm}0.09$
  & $22.95{\pm}0.34$ & $6.21{\pm}0.08$ \\
w/o \Lim
  & $21.84{\pm}0.54$ & $7.12{\pm}0.12$
  & $28.37{\pm}0.68$ & $5.84{\pm}0.11$ \\
w/o \Lan
  & $10.54{\pm}0.14$ & $9.11{\pm}0.06$
  & $12.38{\pm}0.19$ & $7.28{\pm}0.05$ \\
TIRT: fixed schedule
  & $11.47{\pm}0.17$ & $8.62{\pm}0.07$
  & $13.14{\pm}0.21$ & $7.03{\pm}0.06$ \\
TIRT: learned $\lambda_0{=}0$
  & $12.71{\pm}0.19$ & $8.44{\pm}0.07$
  & $14.41{\pm}0.24$ & $6.88{\pm}0.06$ \\
\dash~(ours)
  & $\mathbf{8.87{\pm}0.12}$ & $\mathbf{9.31{\pm}0.06}$
  & $\mathbf{10.47{\pm}0.16}$ & $\mathbf{7.41{\pm}0.05}$ \\
\bottomrule
\end{tabular}
\end{table}



\subsection{Calibration Views}
\label{app:calibration_views}

Figure~\ref{fig:calibration_views} reports the per-timestep and
training-dynamics views summarised in the main text.
Panel~(a): \dash maintains $\rho(t){\in}[0.80,\,0.96]$ across the
denoising trajectory, with strongest preservation in the
structure-forming regime ($t{>}700$); composite collapses
near-uniformly to $\rho(t){\approx}0.68$ and gap matching shows
intermediate behaviour ($\rho(t){\in}[0.75,\,0.88]$), improving over
composite but unable to anchor individual branches. Panel~(b):
\dash stabilises at $\rho{=}0.91$ by
epoch~100, while w/o~\Lun decays from near~1.0 to 0.23;
$\rho{\approx}1$ at initialisation reflects an untrained gap of
incidental magnitude rather than calibration.
Panel~(c): distilled configurations cluster by supervision type in the
$(\rho,\cos(\Delta))$ plane, the teacher-free reference omitted, with \dash at (0.91,~0.94) and
single-constraint baselines toward the lower left. Because the
per-sample ratio $\|\Delta_S\|/\|\Delta_T\|$ is tightly concentrated
for \dash, w/o~\Lun, w/o~\Lan, both TIRT
variants, composite, gap matching and conditional-only (which enters
with $\cos(\Delta){\approx}0$, its gap being both tiny and
directionless), $\mathrm{MSE}_\Delta$
closely tracks $\rho$ and $\cos(\Delta)$ there: with
$\mathbb{E}[\tfrac{1}{d}\|\Delta_T\|^2]$ measured at $0.240$, $0.289$
and $0.241$ on the three datasets, the aggregate approximation
$\mathbb{E}[\tfrac{1}{d}\|\Delta_T\|^2](1{+}\rho^2{-}2\rho\cos(\Delta))$ tracks
those rows within $2\%$ on CIFAR-10 and $5\%$ on CIFAR-100. The relation
is exact per sample; in aggregate it replaces per-sample ratios by
$\rho$ and $\cos(\Delta)$ and so discards their covariance. FitNets sits inside the same band
($-1.3\%$ and $-1.7\%$) despite supervising features rather than
branches. The approximation degrades where the ratio is heteroscedastic:
label-dropout, magnitude pruning and w/o~\Lim{} deviate by $-21\%$,
$-38\%$ and $+48\%$ on CIFAR-10 and $-35\%$, $-40\%$ and $+41\%$ on
CIFAR-100, multi-$w$ composite by $+8\%$ and $+18\%$, and the teacher-free
\textit{Scratch} reference by $+8\%$ and $-1\%$. The three quantities are therefore not algebraically
determined by one another in general, only empirically
near-determined where both branches are supervised directly.

\begin{figure}[t]
\centering
\setlength{\tabcolsep}{3pt}
\resizebox{\textwidth}{!}{%
\begin{tabular}{@{}ccc@{}}
\figorbox{0.30\textwidth}{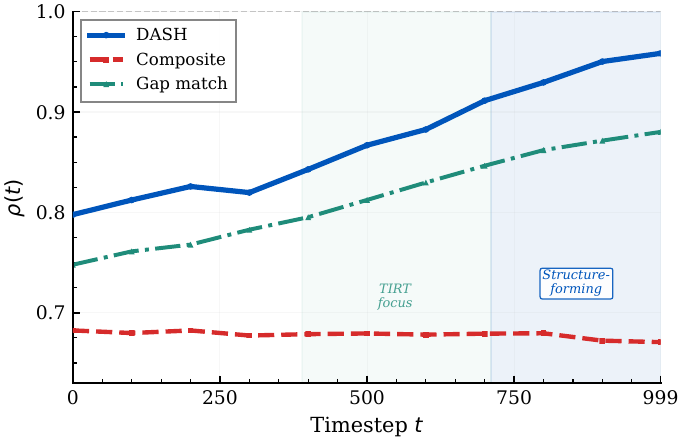} &
\figorbox{0.30\textwidth}{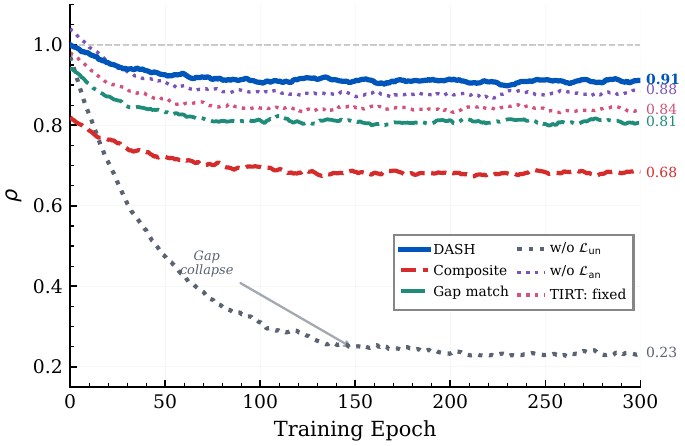} &
\figorbox{0.30\textwidth}{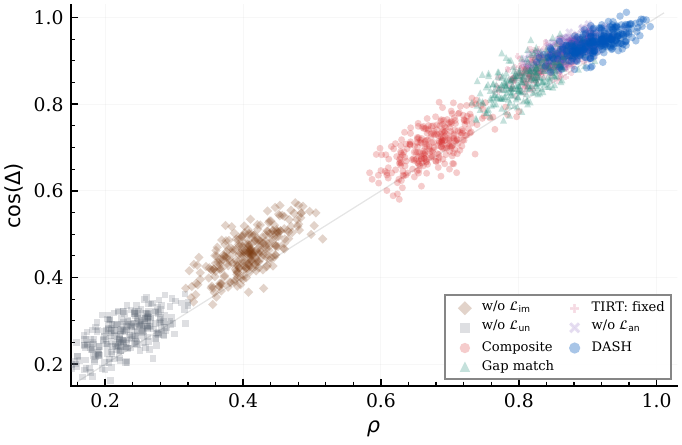} \\[2mm]
{\small (a)~Per-timestep $\rho(t)$} & {\small (b)~Training dynamics} &
{\small (c)~$(\rho,\cos(\Delta))$ plane}
\end{tabular}}
\caption{Calibration views, CIFAR-10.
(a)~per-timestep $\rho(t)$; (b)~$\rho$ over training;
(c)~configurations in the
$(\rho,\cos(\Delta))$ plane. Tabulated values in Tables~1 and~3.}
\label{fig:calibration_views}
\end{figure}

\subsection{Step-Count Ablation}
\label{app:steps}

The student is trained at 50 steps; Table~\ref{tab:steps} evaluates
inference at 5, 10, 25, and 50 steps without retraining. The
teacher--student FID gap remains stable at 3.4--4.4 points on
CIFAR-10, confirming that dual-branch supervision introduces no
additional step-sensitivity beyond the inherent capacity gap. At 5
steps both models degrade by comparable margins ($+$3.65 teacher,
$+$4.68 student on CIFAR-10), indicating that degradation at very
few steps reflects DDIM discretisation error rather than distillation
quality. CIFAR-100 follows the same trend, with the gap growing
modestly from 3.67 to 4.20 FID between 50 and 5 steps.
The C100 gap is non-monotonic across intermediate step counts
($4.20{\to}4.06{\to}4.20$), reflecting that DDIM discretisation
error need not vary monotonically with step count. ImageNet-64 holds
the tightest gap of the three settings at every step count
($2.76$--$3.65$), so the calibration advantage does not erode as the
sampling budget is reduced at 1000 classes.

\begin{table}[ht]
\caption{Step-count ablation. FID$\downarrow$, 50K samples, TAG
$w_{\max}{=}4.0$. The student is trained at 50 steps; the other counts
are inference-time only. \emph{Gap} is student minus teacher at the
same step count. Means over the same three seeds as Table~\ref{tab:distill_std}.}
\label{tab:steps}
\centering\small\setlength{\tabcolsep}{5pt}
\begin{tabular}{llcccc}
\toprule
Dataset & Model & 50 & 25 & 10 & 5 \\
\midrule
\multirow{3}{*}{C10}
  & Teacher      & 5.47 & 5.61 & 6.84  & 9.12  \\
  & \dash~ch=64  & 8.87 & 9.25 & 10.44 & 13.55 \\
  & Gap          & 3.40 & 3.64 & 3.60  & 4.43  \\
\midrule
\multirow{3}{*}{C100}
  & Teacher      & 6.80  & 7.03  & 8.62  & 11.74 \\
  & \dash~ch=64  & 10.47 & 11.23 & 12.68 & 15.94 \\
  & Gap          & 3.67  & 4.20  & 4.06  & 4.20  \\
\midrule
\multirow{3}{*}{IN64}
  & Teacher      & 6.34 & 6.75  & 8.23  & 10.56 \\
  & \dash~ch=64  & 9.10 & 9.96  & 11.10 & 14.21 \\
  & Gap          & 2.76 & 3.21  & 2.87  & 3.65  \\
\bottomrule
\end{tabular}
\end{table}

\subsection{Standard (non-TIRT) Teacher}
\label{app:std_teacher}

Table~\ref{tab:stdteacher} repeats the core comparison with a teacher
trained under standard Min-SNR weighting without TIRT, evaluated at
fixed $w{=}1.5$ without TAG. The
\dash advantage over composite distillation is larger with this
weaker teacher (9.58 FID points at fixed $w{=}1.5$) than with the TIRT
teacher (4.97 points under TAG), so the advantage is not specific to
the TIRT/TAG recipe.

\begin{table}[t]
\caption{Standard Min-SNR teacher, trained without TIRT and evaluated
without TAG, CIFAR-10, single seed. FID$\downarrow$\,/\,IS$\uparrow$, 50K samples, 50-step DDIM,
fixed $w{=}1.5$ for every row; the teacher value is the Min-SNR base of
Table~\ref{tab:teacher}, where the same teacher reaches 7.42 under TAG.}
\label{tab:stdteacher}
\centering\small\setlength{\tabcolsep}{3pt}
\begin{tabular}{lcccccc}
\toprule
Model & Params & FID$\downarrow$ & IS$\uparrow$ & $\rho$
      & $\cos(\Delta)$ & $\mathrm{MSE}_\Delta$ \\
\midrule
Teacher      & 35.8M & 9.21  & 8.54 & 1.00 & 1.00 & 0.000 \\
Composite    & 6.1M  & 23.32 & 7.65 & 0.53 & 0.58 & 0.214 \\
\dash~(ours) & 6.1M  & \textbf{13.74} & \textbf{8.21}
             & \textbf{0.81} & \textbf{0.84} & \textbf{0.080} \\
\bottomrule
\end{tabular}
\end{table}

\subsection{Width Scaling}
\label{app:width}

Table~\ref{tab:width} evaluates ch=48, 64, and 96 under identical
conditions. Distillation gain over scratch is consistent across the
$2.5{\times}$--$10.5{\times}$ compression range, and $\rho$ scales
monotonically ($0.86{\to}0.91{\to}0.94$), remaining well above the
composite baseline measured at ch=64 ($\rho{=}0.68$) at every width.
The teacher-free reference does not follow it: \textit{Scratch} holds
$\rho$ near $0.81$, $0.80$ and $0.83$ across the same three widths
while its $\cos(\Delta)$ rises from $0.45$ to $0.58$. Capacity therefore raises gap
magnitude only under distillation, while direction improves with width
either way. At $10.5{\times}$
compression the student achieves FID~12.24 against scratch FID~22.14,
confirming that dual-branch supervision is effective even under
severe capacity constraint. Figure~\ref{fig:pareto} visualises this
trade-off directly.

\begin{figure}[t]
\centering
\figorbox{0.50\linewidth}{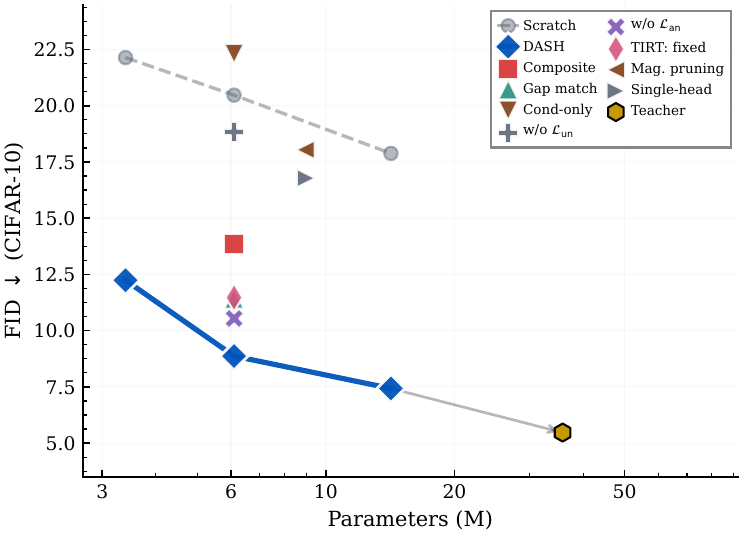}
\caption{Compression--quality Pareto frontier (CIFAR-10). \dash
leads across $2.5{\times}$--$10.5{\times}$; magnitude pruning and
single-head carry more parameters and still fall short. Values in
Tables~1 and~3.}
\label{fig:pareto}
\end{figure}

\begin{table}[t]
\caption{Width scaling on CIFAR. FID$\downarrow$\,/\,IS$\uparrow$,
50-step DDIM, TAG $w_{\max}{=}4.0$, 234K iterations. Calibration on CIFAR-10.
Widths are relative to the CIFAR teacher (ch=128,
$\mathrm{nrb}{=}2$). The ch=64 rows reproduce the three-seed means of
Table~\ref{tab:distill_std}.}
\label{tab:width}
\centering\small\setlength{\tabcolsep}{4pt}
\begin{tabular}{lcccccccc}
\toprule
& & & \multicolumn{2}{c}{C10} & \multicolumn{2}{c}{C100}
& \multicolumn{2}{c}{Calib.\ (C10)} \\
\cmidrule(lr){4-5}\cmidrule(lr){6-7}\cmidrule(lr){8-9}
Config & Params & Comp.
& FID$\downarrow$ & IS$\uparrow$
& FID$\downarrow$ & IS$\uparrow$
& $\rho$ & $\cos(\Delta)$ \\
\midrule
Teacher & 35.8M & $1.0{\times}$ & 5.47 & 9.42 & 6.80 & 7.46
  & 1.00 & 1.00 \\
\midrule
Scratch ch=48 & 3.4M  & $10.5{\times}$ & 22.14 & 7.11 & 28.79 & 6.22
  & 0.81 & 0.45 \\
\dash~ch=48   & 3.4M  & $10.5{\times}$ & 12.24 & 8.88 & 15.77 & 7.12
  & 0.86 & 0.91 \\
\midrule
Scratch ch=64 & 6.1M  & $5.9{\times}$  & 20.47 & 7.44 & 26.14 & 6.51
  & 0.80 & 0.51 \\
\dash~ch=64   & 6.1M  & $5.9{\times}$  &
  $\mathbf{8.87}$ & $\mathbf{9.31}$ & $\mathbf{10.47}$ & $\mathbf{7.41}$
  & 0.91 & 0.94 \\
\midrule
Scratch ch=96 & 14.2M & $2.5{\times}$  & 17.88 & 7.68 & 22.08 & 6.74
  & 0.83 & 0.58 \\
\dash~ch=96   & 14.2M & $2.5{\times}$  &  7.43 & 9.38 &  8.91 & 7.44
  & 0.94 & 0.96 \\
\bottomrule
\end{tabular}
\end{table}

\subsection{Guidance Schedule Generalisation}
\label{app:guidance}

Table~\ref{tab:guidance} evaluates FID across
$w_{\max}{\in}\{1.5,\ldots,5.0\}$. Both teacher and student follow the
expected U-shaped curve with minimum near TAG $w_{\max}{=}4.0$. The
teacher--student gap is U-shaped, with a minimum of 2.89 points at
$w_{\max}{=}3.0$ and widening at both tails to 3.45 at
$w_{\max}{=}1.5$ and 3.88 at $w_{\max}{=}5.0$. The calibration ratio $\rho{=}0.91$ is measured on branch outputs and
is therefore independent of the guidance schedule.

\begin{table}[t]
\caption{Guidance scale generalisation. FID$\downarrow$, CIFAR-10,
ch=64, 50K samples, 50 steps. Column headings are TAG $w_{\max}$; the bold
column marks the evaluation setting used throughout the paper. Neither
model is trained at a fixed guidance weight. TAG holds
$w_{\min}{=}1.0$ at every $w_{\max}$, so the leftmost column spans only
$[1.04,1.46]$ and is effectively unguided across the high-noise half of
the trajectory; it is not comparable to a fixed $w{=}1.5$ run, which
guides at every timestep. Robustness to the
inference-time schedule is a property of the distillation objective and
is demonstrated on CIFAR-10. Means over the same three seeds as Table~\ref{tab:distill_std}.}
\label{tab:guidance}
\centering\small\setlength{\tabcolsep}{12pt}
\resizebox{\textwidth}{!}{%
\begin{tabular}{lcccccccc}
\toprule
Model & $1.5$ & $2.0$ & $2.5$ & $3.0$ & $3.5$ & $\mathbf{4.0}$
      & $4.5$ & $5.0$ \\
\midrule
Teacher
  & 10.42 & 8.31 & 7.14 & 6.52 & 5.89 & \textbf{5.47} & 5.61 & 6.23 \\
\dash~ch=64
  & 13.87 & 11.38 & 10.12 & 9.41 & 8.94 & \textbf{8.87} & 9.23 & 10.11 \\
Gap
  & 3.45 & 3.07 & 2.98 & 2.89 & 3.05 & \textbf{3.40} & 3.62 & 3.88 \\
\bottomrule
\end{tabular}}
\end{table}

\subsection{Fixed-\texorpdfstring{$w$}{w} Versus Shared TAG}
\label{app:protocol}
Every result in this paper uses the shared TAG schedule at evaluation.
Table~\ref{tab:protocol} repeats the CIFAR-10 comparison under a fixed
guidance weight instead, holding the checkpoints fixed. The ordering is
unchanged, so the branch-level conclusion does not depend on the
guidance schedule. The margins in fact widen: \dash leads composite
distillation by 4.97 FID under TAG and 5.56 under fixed guidance, and
the ablation without \Lun by 9.97 and 10.67. The FID cost of fixed
guidance grows as calibration falls ($0.63$ for the teacher at
$\rho{=}1.00$, $0.78$ for \dash at $0.91$, $1.37$ for composite at
$0.68$, $1.48$ without \Lun at $0.23$), as $\eps^\varnothing_S{+}w\Delta_S$
predicts: a fixed weight amplifies $\Delta_S$ equally at every
timestep, so a poorly calibrated gap is carried further from the
teacher's.

\begin{table}[t]
\caption{Guidance protocol comparison (CIFAR-10, ch=64, 50K samples,
50-step DDIM). Inference only; the same checkpoints are evaluated under
both protocols. The TAG column reproduces the corresponding rows of
the main text's Tables~1 and~3. Means over the same three seeds as Table~\ref{tab:distill_std}.}
\label{tab:protocol}
\centering\small\setlength{\tabcolsep}{4pt}
\begin{tabular}{lcc}
\toprule
Model & FID$\downarrow$, fixed $w{=}4.0$ & FID$\downarrow$, TAG $w_{\max}{=}4.0$ \\
\midrule
Teacher      & 6.10  & 5.47 \\
Composite    & 15.21 & 13.84 \\
w/o \Lun     & 20.32 & 18.84 \\
\midrule
\dash~(ours) & \textbf{9.65} & \textbf{8.87} \\
\bottomrule
\end{tabular}
\end{table}


\section{Qualitative Results}
\label{app:qualitative}

Figures are grouped by type: class-conditional diversity
(\S\ref{app:qual_diversity}), sampling-step behaviour
(\S\ref{app:qual_steps}), and failure cases
(\S\ref{app:qual_failure}). All grids use 50-step DDIM at
TAG $w_{\max}{=}4.0$ with seeds matched between teacher and student.

\subsection{Class-Conditional Diversity}
\label{app:qual_diversity}

Each grid is $5{\times}10$: columns are classes, rows are five
independent samples per class. Teacher and \dash student appear as
stacked panels within a
single figure per dataset, for direct comparison: CIFAR-10
(Figure~\ref{fig:teacher_c10_div}), CIFAR-100
(Figure~\ref{fig:teacher_c100_div}), and ImageNet-64
(Figure~\ref{fig:teacher_in64_div}).

Collapse drives samples toward the unconditional distribution, so the
comparison to make is class fidelity within each labelled column and
separation between columns, not diversity alone: strong guidance in
fact \emph{reduces} within-column variety. The
student's columns retain the teacher's class identity and spread on
all three datasets.

\begin{figure}[t]
\centering
\figorbox{\linewidth}{cifar10_teacher_5x10.pdf}\\[0.5mm]
{\small (a)~Teacher (35.8M, FID~5.47, IS~9.42)}\\[2mm]
\figorbox{\linewidth}{cifar10_student_5x10.pdf}\\[0.5mm]
{\small (b)~\dash student (6.1M, FID~8.87, IS~9.31, $\rho{=}0.91$)}
\caption{\textbf{CIFAR-10 class-conditional diversity}, same classes
and seeds throughout; columns left to right: Plane, Car, Bird, Cat,
Deer, Dog, Frog, Horse, Ship, Truck. Class identity and colour
composition are largely preserved at $5.9{\times}$ compression, with the
residual gap appearing as texture softening rather than class
confusion.}
\label{fig:teacher_c10_div}
\end{figure}

\begin{figure}[t]
\centering
\figorbox{\linewidth}{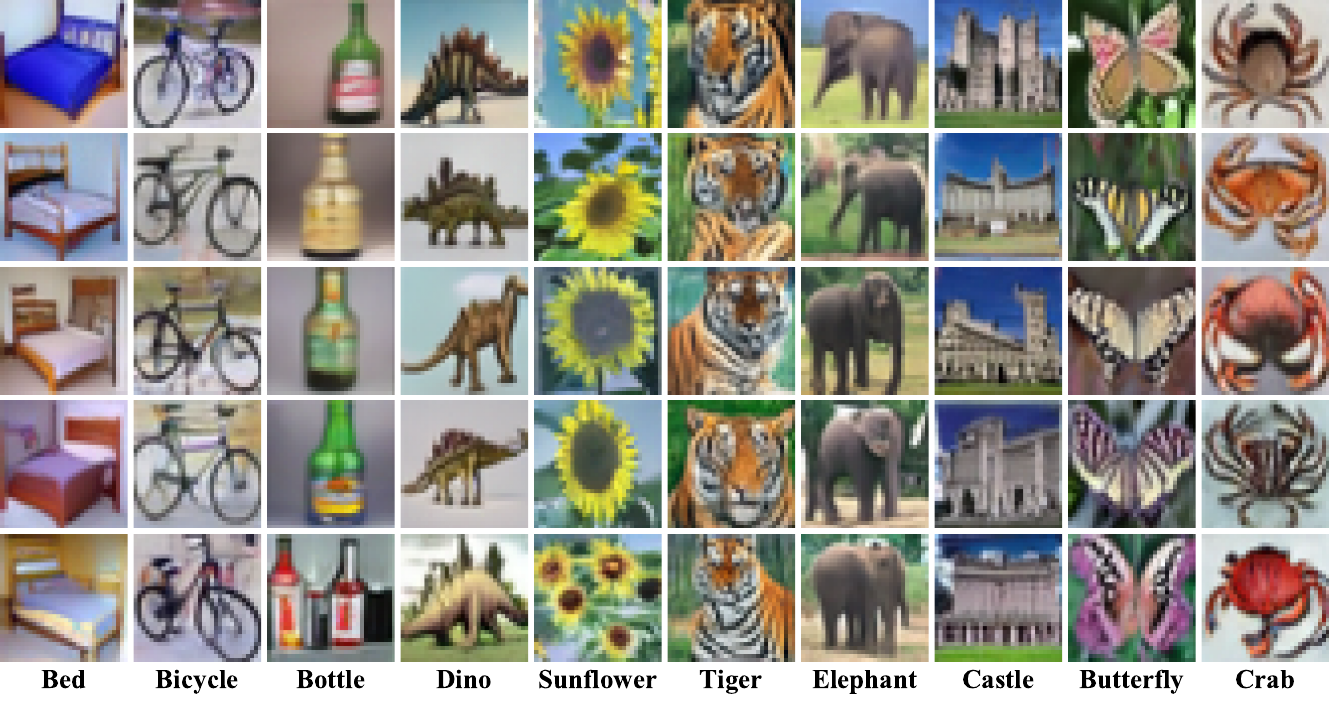}\\[0.5mm]
{\small (a)~Teacher (35.8M, FID~6.80, IS~7.46)}\\[2mm]
\figorbox{\linewidth}{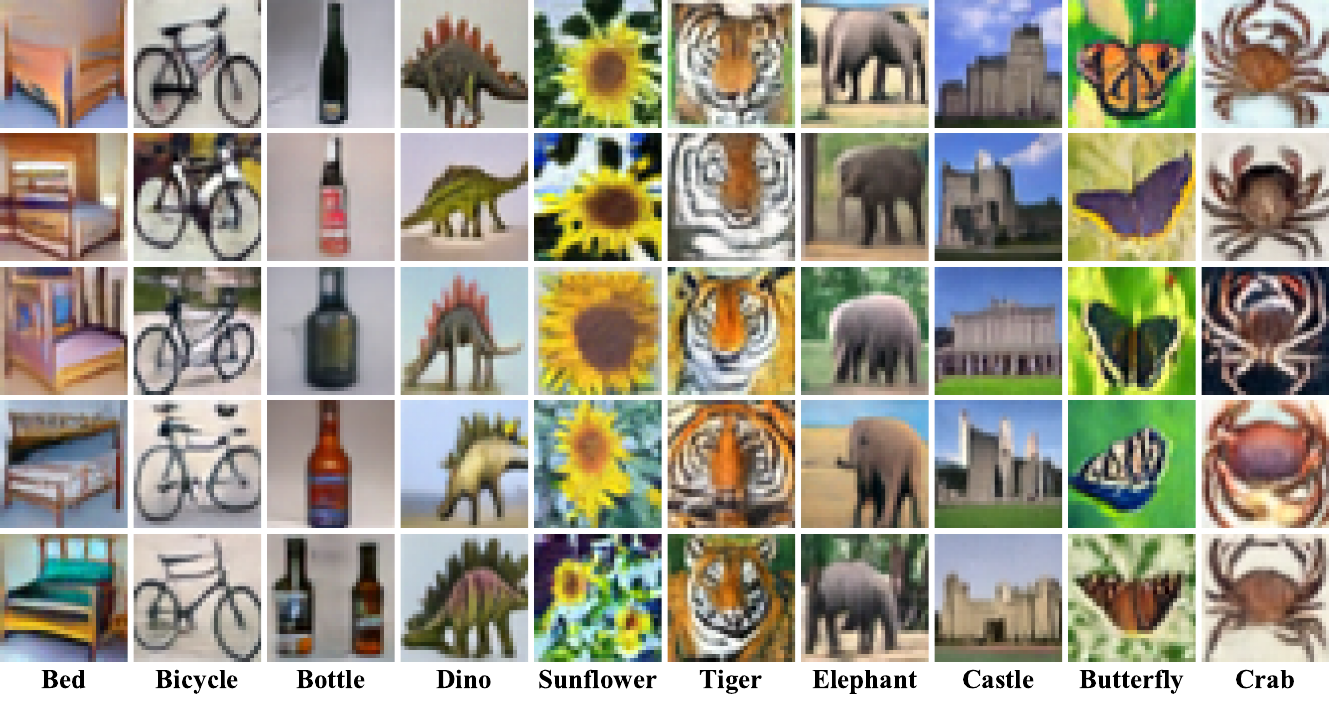}\\[0.5mm]
{\small (b)~\dash student (6.1M, FID~10.47, IS~7.41, $\rho{=}0.89$)}
\caption{\textbf{CIFAR-100 class-conditional diversity}, same classes
and seeds throughout; columns left to right: Bed, Bicycle, Bottle,
Dinosaur, Sunflower, Tiger, Elephant, Castle, Butterfly, Crab.
Degradation is more visible than on CIFAR-10, consistent with the
larger FID gap over 100 fine-grained classes.}
\label{fig:teacher_c100_div}
\end{figure}

\begin{figure}[t]
\centering
\figorbox{\linewidth}{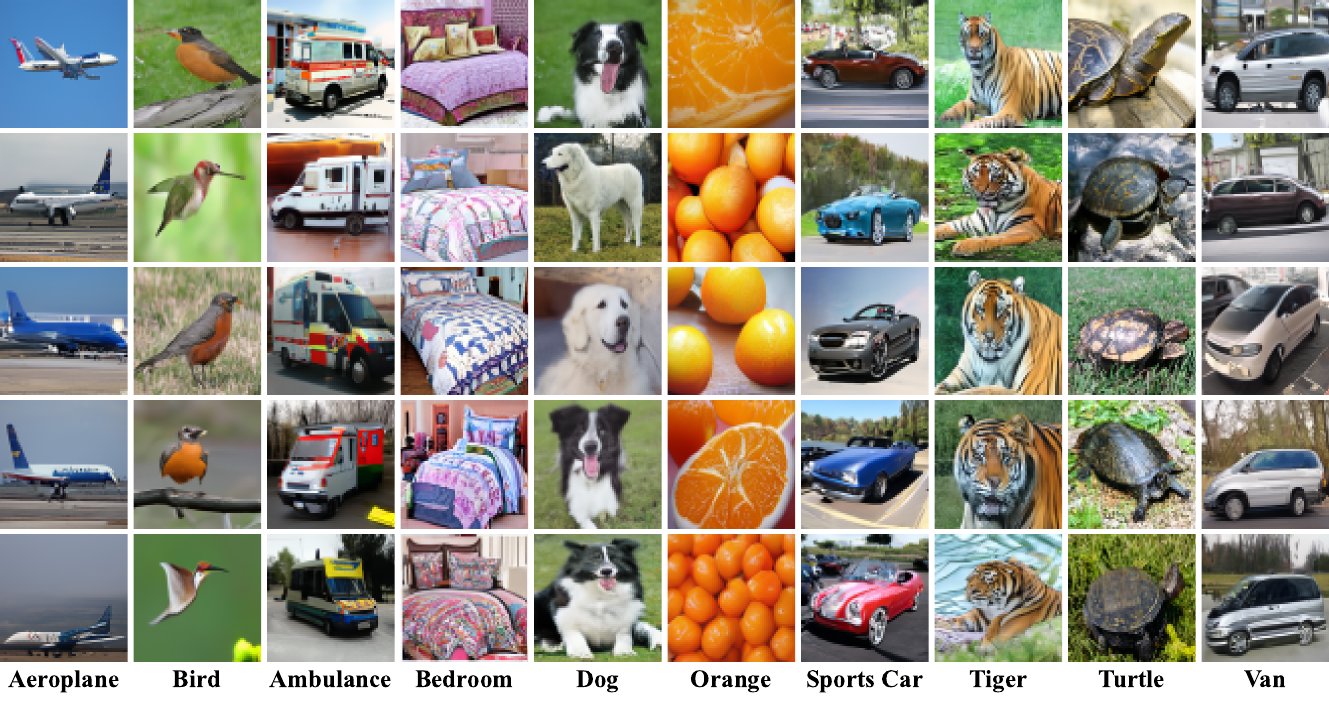}\\[0.5mm]
{\small (a)~Teacher (116.5M, FID~6.34, IS~48.35)}\\[2mm]
\figorbox{\linewidth}{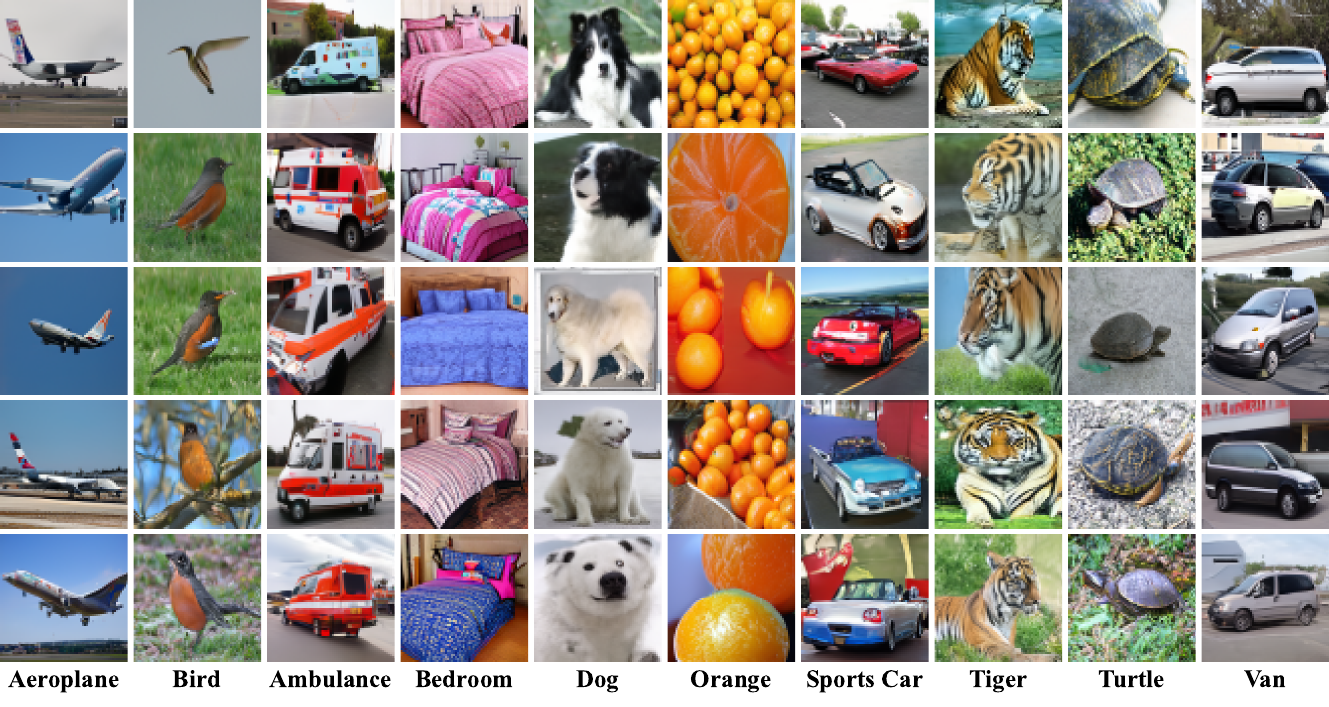}\\[0.5mm]
{\small (b)~\dash student (22.3M, $5.2{\times}$, FID~9.10, IS~47.32,
$\rho{=}0.89$)}
\caption{\textbf{ImageNet-64 class-conditional diversity}, same
classes and seeds throughout, drawn from the 1000-class label set;
columns as labelled. TAG $w_{\max}{=}4.0$, 50-step DDIM, matching the
protocol used for the CIFAR grids. Class identity and intra-class
variation are largely preserved: at $\rho{=}0.89$ the student retains
most of the teacher's guidance gap, so the residual deficit appears primarily
as softened texture rather than as the class confusion that branch
collapse produces. This matches the quantitative
result in the main text's Table~2.}
\label{fig:teacher_in64_div}
\end{figure}

\subsection{Sampling-Step Comparison}
\label{app:qual_steps}

Figure~\ref{fig:steps_all} shows the same student checkpoint sampled at
50, 25, 10 and 5 DDIM steps on all three datasets. Structure and class
identity survive step
reduction well below the 50-step operating point; degradation appears
first as high-frequency detail loss, not as a change in what the sample
depicts.

\begin{figure}[t]
\centering
\setlength{\tabcolsep}{2pt}
\resizebox{\textwidth}{!}{%
\begin{tabular}{@{}c cccc@{}}
& {\small 50 steps} & {\small 25 steps} & {\small 10 steps} & {\small 5 steps} \\[0.5mm]
\rotatebox{90}{\small CIFAR-10} &
\figorbox{0.225\textwidth}{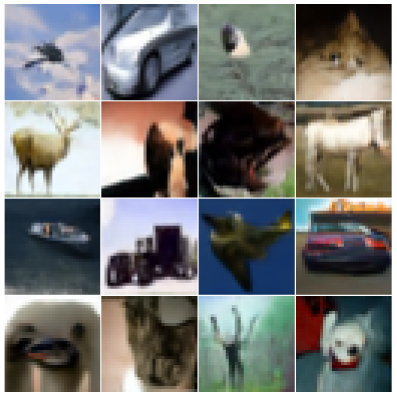} &
\figorbox{0.225\textwidth}{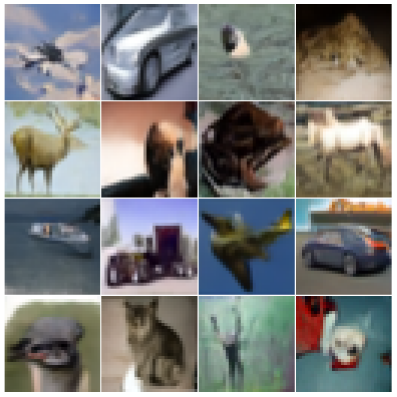} &
\figorbox{0.225\textwidth}{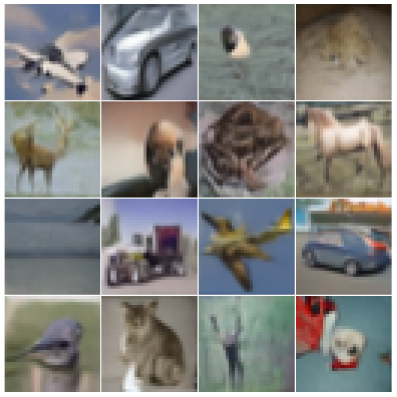} &
\figorbox{0.225\textwidth}{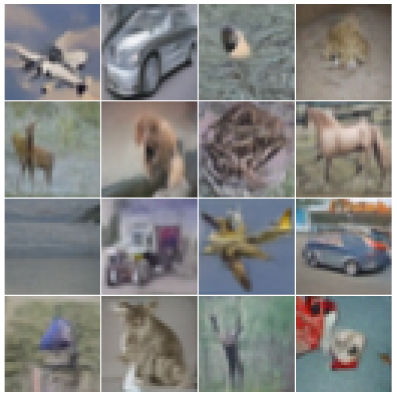} \\[0.3mm]
& {\scriptsize FID~8.87} & {\scriptsize FID~9.25}
& {\scriptsize FID~10.44} & {\scriptsize FID~13.55} \\[1.5mm]
\rotatebox{90}{\small CIFAR-100} &
\figorbox{0.225\textwidth}{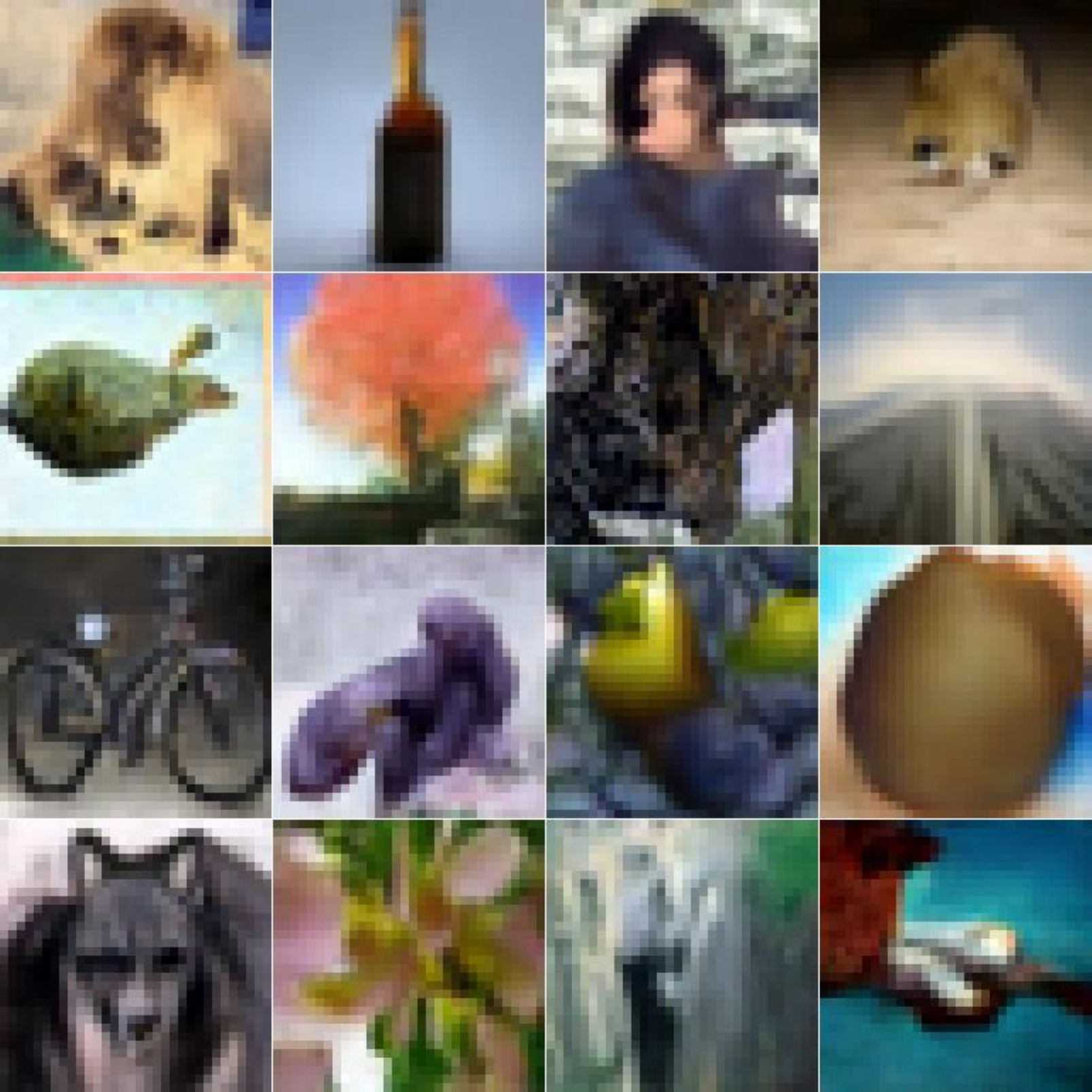} &
\figorbox{0.225\textwidth}{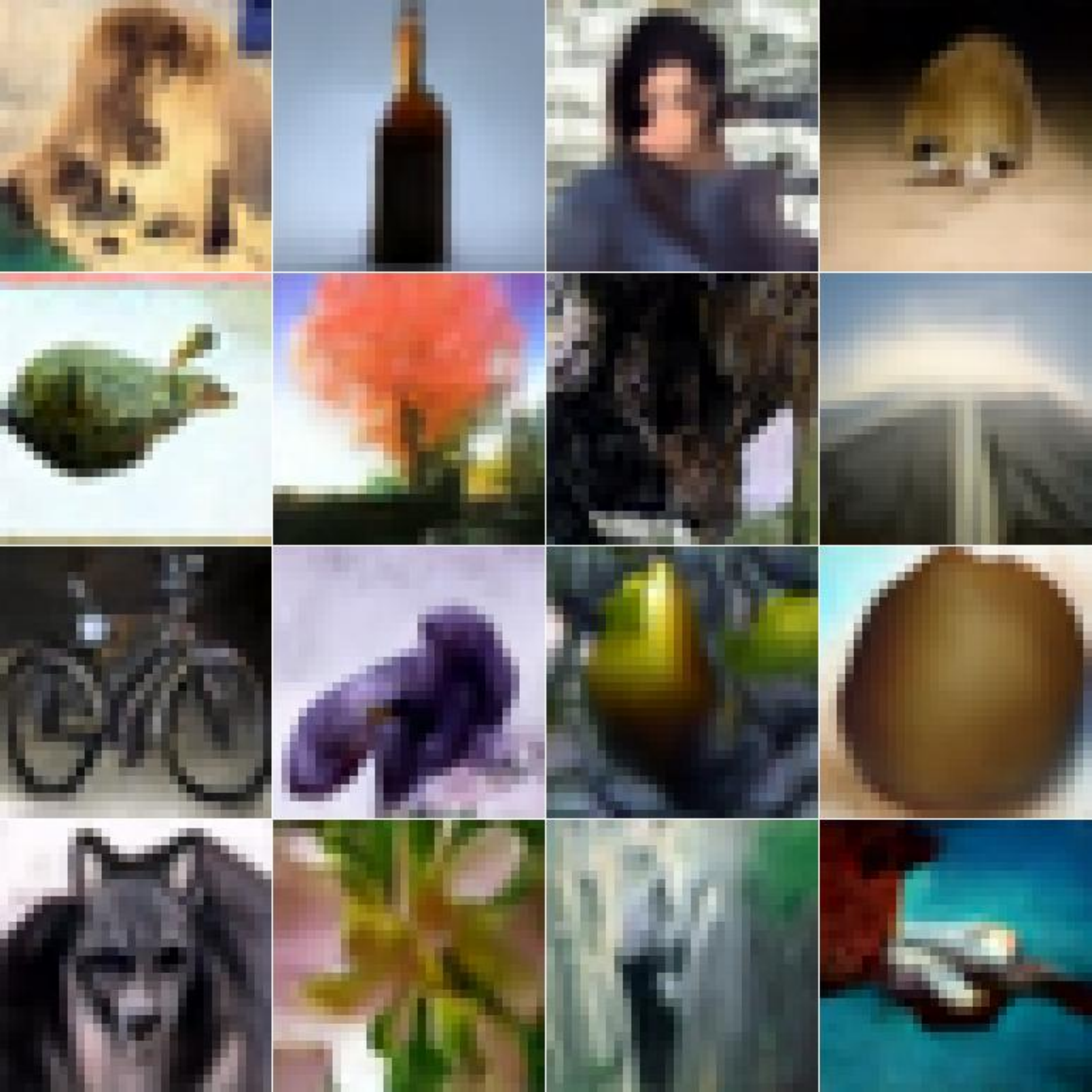} &
\figorbox{0.225\textwidth}{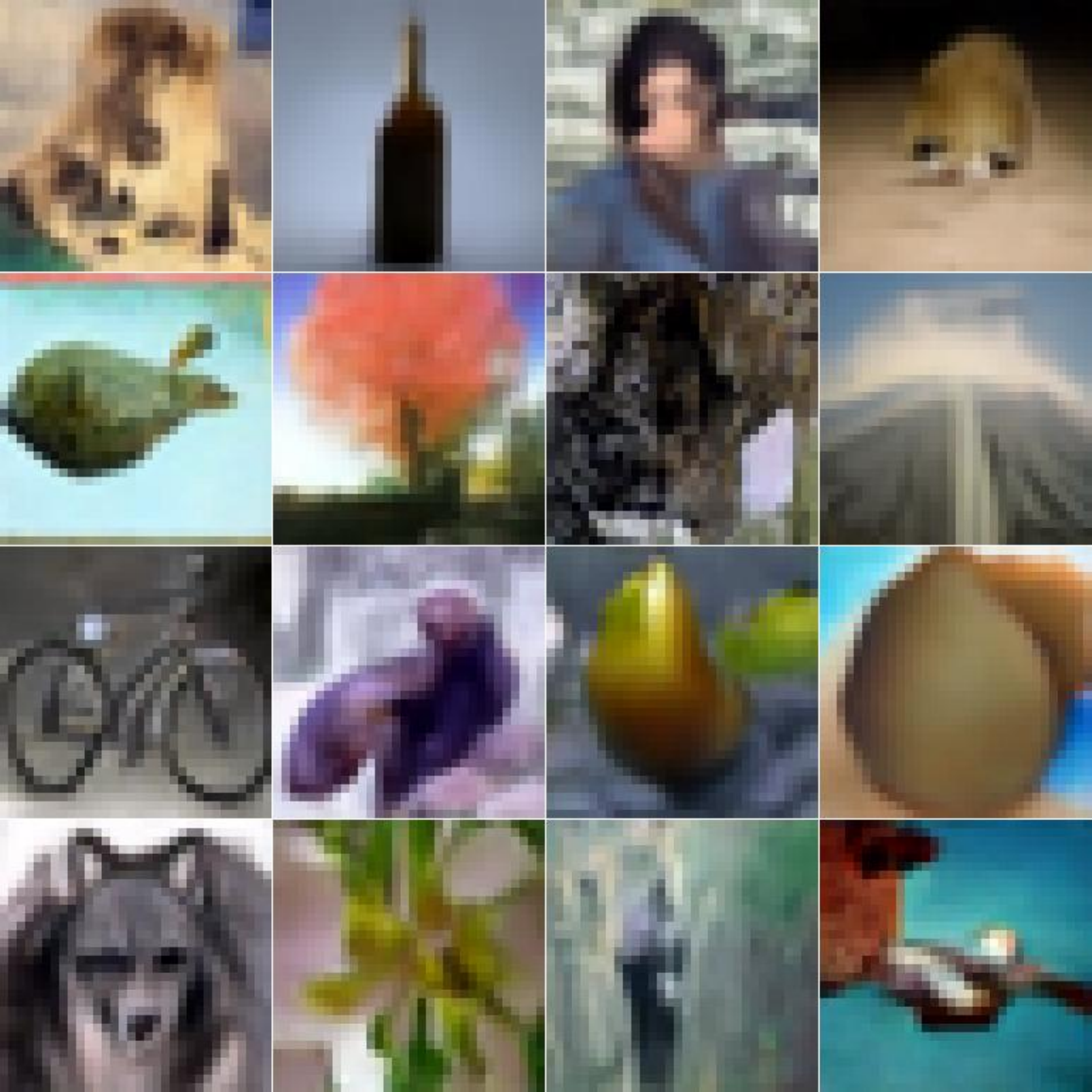} &
\figorbox{0.225\textwidth}{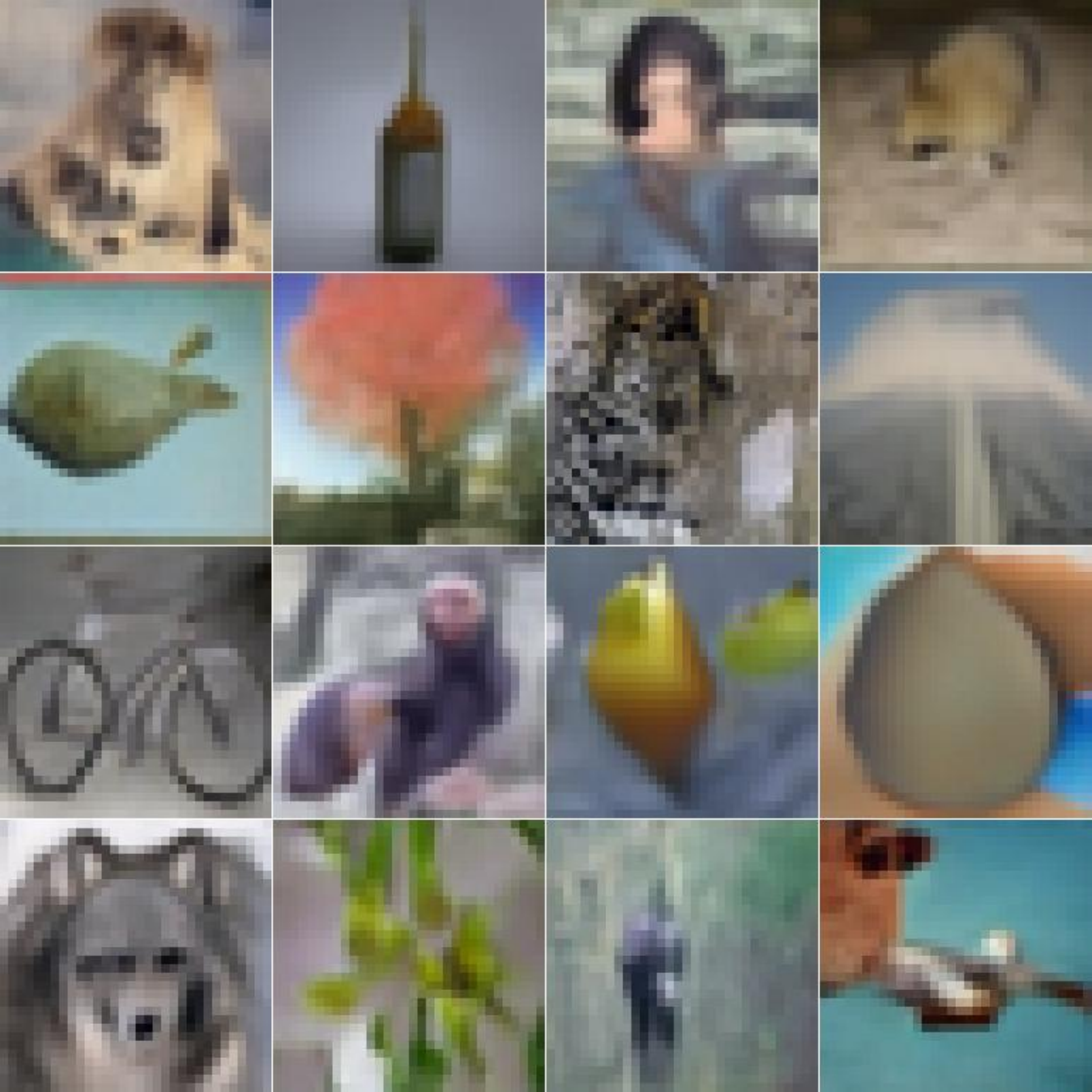} \\[0.3mm]
& {\scriptsize FID~10.47} & {\scriptsize FID~11.23}
& {\scriptsize FID~12.68} & {\scriptsize FID~15.94} \\[1.5mm]
\rotatebox{90}{\small ImageNet-64} &
\figorbox{0.225\textwidth}{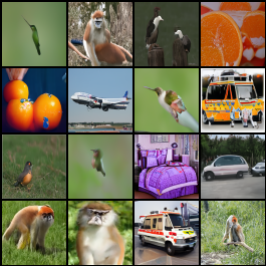} &
\figorbox{0.225\textwidth}{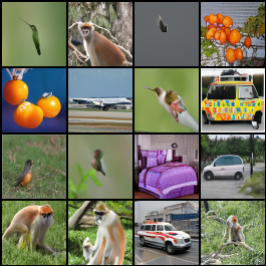} &
\figorbox{0.225\textwidth}{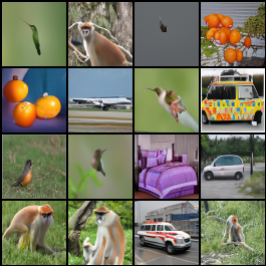} &
\figorbox{0.225\textwidth}{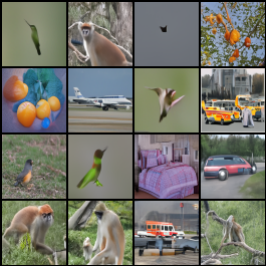} \\[0.3mm]
& {\scriptsize FID~9.10} & {\scriptsize FID~9.96}
& {\scriptsize FID~11.10} & {\scriptsize FID~14.21}
\end{tabular}}
\caption{Step-count comparison for the \dash student across all three
datasets. Columns are 50, 25, 10 and 5 DDIM steps; rows are CIFAR-10,
CIFAR-100 and ImageNet-64. Degradation is gradual to 10 steps and
becomes visible at 5.}
\label{fig:steps_all}
\end{figure}

\subsection{Failure Cases}
\label{app:qual_failure}

Worst-case student outputs, selected by largest per-sample
guidance-gap error against the teacher
(Figure~\ref{fig:failure}). Selection is on the calibration criterion
rather than FID, so these are the cases where the gap departs
furthest from the teacher's. Under branch collapse the worst cases
would drift toward the unconditional distribution; the degradation
here is confined to fine texture, pointing to a capacity limit
instead.

\begin{figure}[t]
\centering
\figorbox{\linewidth}{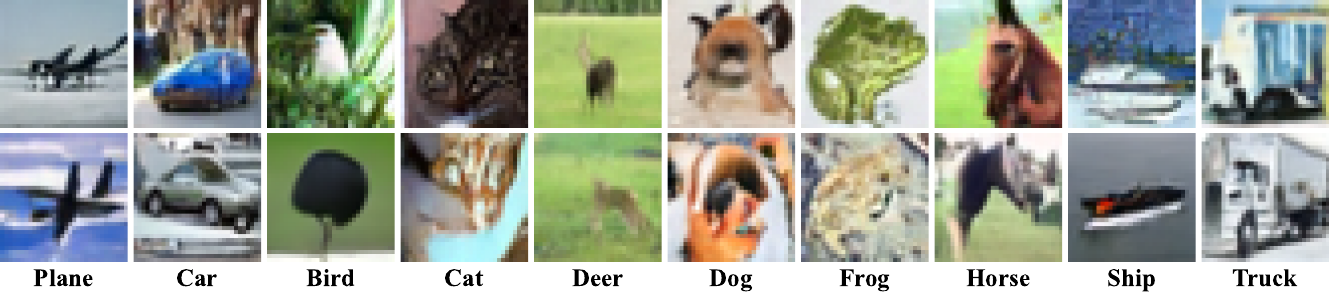}\\[0.5mm]
{\small (a)~CIFAR-10}\\[2mm]
\figorbox{\linewidth}{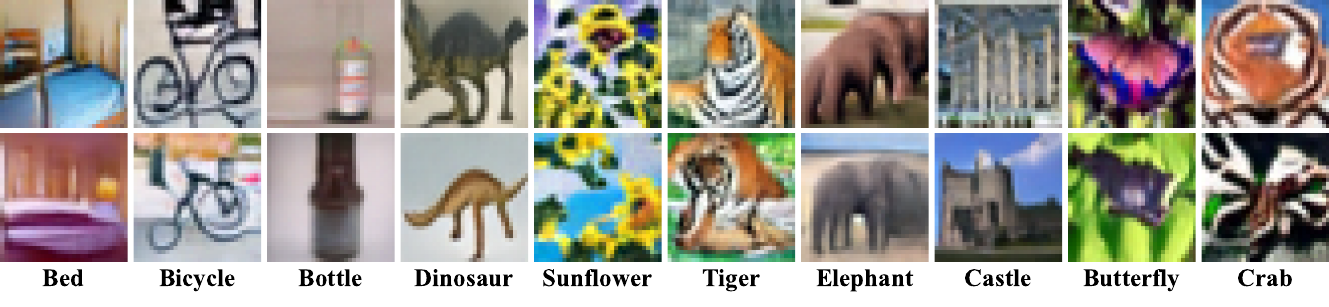}\\[0.5mm]
{\small (b)~CIFAR-100}\\[2mm]
\figorbox{\linewidth}{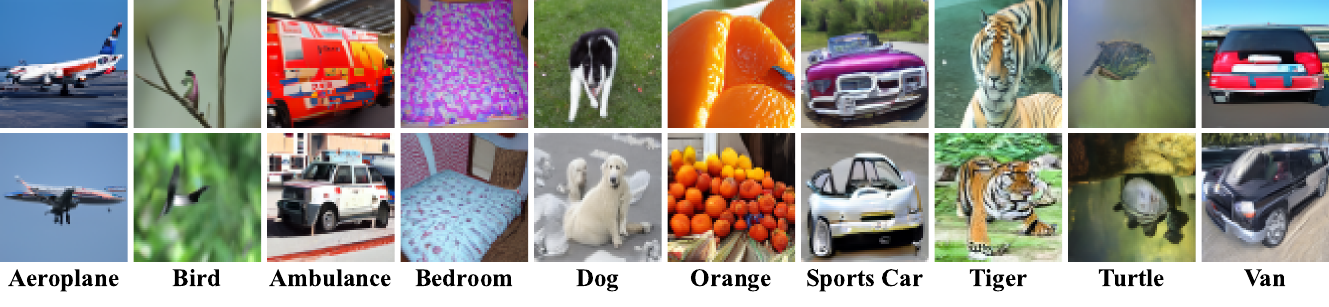}\\[0.5mm]
{\small (c)~ImageNet-64}
\caption{Worst-case \dash student outputs; top row teacher, bottom row
\dash student. Texture softening and loss of fine structure dominate,
and class identity is largely retained; failures concentrate in
fine-grained categories whose distinguishing features are small or
low-contrast, and the same pattern holds at 1000 classes.}
\label{fig:failure}
\end{figure}

\section{Limitations, Broader Impact, and LLM Usage}
\label{app:limitations}

\textbf{Limitations.} The main text states the scale and
teacher-architecture scope of these results; two further boundaries
apply here. TIRT and TAG are evaluated under one teacher recipe.
\S\ref{app:std_teacher} repeats the core comparison with a standard
Min-SNR teacher and the ordering holds, but a single additional recipe
does not establish generality. The width and depth study in
\S\ref{app:width} holds the training budget fixed across
configurations, isolating capacity as the single variable. Parameter compression and step reduction are argued to
be composable in the main text's related work; combining them is left
to future work.

\textbf{Broader impact.} This work lowers the compute cost of guided
diffusion sampling while largely preserving class-conditional fidelity,
easing deployment under constrained hardware. Lower cost may also ease
misuse, a risk inherited from diffusion generation itself rather than
introduced here.

\textbf{Code.} Code is available at
\url{https://github.com/C-loud-Nine/DASH_Dual-Branch-Score-Distillation},
covering CIFAR-10/-100 teacher and student training, all
Table~\ref{tab:distill} baselines, ImageNet-64 training and evaluation, and the
FID\,/\,IS and calibration metrics, with setup instructions.

\textbf{LLM usage.} Large language model tools were used to assist
with editing and accelerate LaTeX formatting.
All technical claims, derivations, and experimental results were
authored, verified, and are the responsibility of the authors.

\end{document}